\documentclass[journal]{IEEEtran}
\pdfoutput=1 
\ifCLASSOPTIONcompsoc
\usepackage[nocompress]{cite}
\else
  \usepackage{cite}
\fi
\usepackage{algorithm}
\usepackage{algpseudocode}
\usepackage{epsfig}
\usepackage{graphicx}
\usepackage{amsfonts}
\usepackage{comment}
\usepackage{caption}
\usepackage{color}
\usepackage{amssymb}
\usepackage{amsmath}
\usepackage{amsthm,mathtools}
\usepackage{multirow}
\usepackage{etoolbox}
\usepackage{balance}
\usepackage{booktabs}
\usepackage{enumitem}
\usepackage[table]{xcolor}
\usepackage{hyperref}
\usepackage{upgreek}
\usepackage{tikz}

\usepackage{caption}
\usepackage{boldline}
\usepackage{makecell}
\setcellgapes{1pt}

\AfterEndEnvironment{lemma}{\noindent\ignorespaces}

\AfterEndEnvironment{definition}{\noindent\ignorespaces}

\AfterEndEnvironment{example}{\noindent\ignorespaces}

\ifCLASSINFOpdf
\else
\fi
\begin{document}

\title{FES: A Fast Efficient Scalable QoS Prediction Framework}

\author{Soumi~Chattopadhyay,~\IEEEmembership{Member,~IEEE,}
        Chandranath~Adak,~\IEEEmembership{Member,~IEEE}, and
        Ranjana~Roy~Chowdhury

\IEEEcompsocitemizethanks{\IEEEcompsocthanksitem 
This work was supported by the Science and Engineering Research Board, Department of Science and Technology, Government of India, under Grant SRG/2020/001454.

S. Chattopadhyay is with Department of CSE, Indian Institute of Information Technology Guwahati, India-781015.  
e-mail: $soumi@iiitg.ac.in$

C. Adak is with Centre for Data Science, JIS Institute of Advanced Studies and Research, JIS University, Kolkata, India-700091.


R. R. Chowdhury is with Department of CSE, Indian Institute of Technology Ropar, India-140001.

This work has been submitted to the IEEE for possible publication. Copyright may be transferred without notice, after which this version may no longer be accessible.
}}

\makeatletter
\long\def\@IEEEtitleabstractindextextbox#1{\parbox{0.922\textwidth}{#1}}
\makeatother

\maketitle

\begin{abstract}
Quality-of-Service (QoS) prediction of web service is an integral part of services computing due to its diverse applications in service composition/selection/recommendation. One of the primary objectives of designing a QoS prediction algorithm is to achieve satisfactory prediction accuracy. However, accuracy is not the only criteria to meet while developing a QoS prediction algorithm. The algorithm should be faster in terms of prediction time to be compatible with a real-time system. The other important factor to consider is scalability to tackle large-scale datasets. The existing QoS prediction algorithms often satisfy one goal while compromising the others. In this paper, we propose a semi-offline QoS prediction framework to achieve three important goals together: higher accuracy, faster prediction time, and scalability. Here, we aim to predict the QoS value of service that varies across users. Our framework (FES) consists of multi-phase prediction algorithms: preprocessing-phase prediction, online prediction, and prediction using the proposed pre-trained model. In the preprocessing phase, we first apply multi-level clustering on the dataset to obtain correlated users and services. We then preprocess the clusters using collaborative filtering to remove the sparsity. Finally, we create a two-staged, semi-offline regression model using neural networks to predict the QoS value of service to be invoked by a user in real-time. Our experimental results on WS-DREAM datasets show the efficiency (in terms of accuracy), scalability, and fast responsiveness (in terms of prediction time) of FES as compared to the state-of-the-art methods.
\end{abstract}



\section{Introduction}\label{sec:intro}
\noindent
Web services are becoming ubiquitous with the pervasiveness of smart technology \cite{viejo2020secure}. With the increasing trend of online activities, a substantial number of competing web services facilitating similar functionalities are getting available and thus leading to difficulties in selecting an appropriate service for a specific purpose. Consequently, service recommendation becomes challenging \cite{DBLP:journals/tsc/FanHZWBC21,DBLP:journals/tsc/ShuJYW21}. 

In general, the services are recommended depending on various factors, such as performance \cite{DBLP:journals/fgcs/ChenSYLS20,DBLP:journals/tsc/WangMCYC19}, cost \cite{DBLP:conf/IEEEscc/RamacherM12}, the number of features supported by services  \cite{DBLP:conf/IEEEscc/ChattopadhyayBM16}. Among these distinguishing factors, parameters like cost, the number of features are constant and available prior to the service execution. However, performance-parameters of service vary across different dimensions (e.g., users, locations, time) \cite{DBLP:journals/soca/YuH16,9189762,DBLP:journals/computing/KeshavarziHB20,DBLP:journals/access/LiWLR18}. Therefore, predicting the performance parameters of service plays a pivotal role in service selection. The performance of service is measured by a set of parameters, called Quality-of-Service (QoS) parameters (e.g., response time, throughput, reliability, availability) \cite{zheng2014investigating,DBLP:conf/issre/ZhangZL11}. In this paper, we primarily focus on the variation of the QoS parameters of service across different users by leveraging their location-information.


QoS prediction is a well-studied problem in the literature \cite{DBLP:journals/access/YinZXZMY19,DBLP:conf/icsoc/ZouJNWPG18,DBLP:conf/icws/LeePB15}. The state-of-the-art (SoA) algorithms mainly focus on predicting the QoS value with high accuracy \cite{chowdhury2020cahphf,DBLP:conf/icsoc/ChattopadhyayB19,DBLP:conf/icsoc/ZouJNWPG18,DBLP:journals/tsc/WangMCYC19}. However, considering the need of the hour, the QoS prediction algorithms are also required to be scalable and faster enough in terms of prediction time to be part of real-time systems. Therefore, while designing the QoS prediction algorithms, three important goals are required to be taken into account: (a) higher accuracy, (b) faster prediction, and (c) higher scalability.


Numerous approaches have been adopted for QoS prediction, among which the most widely accepted method is based on collaborative filtering (CF) \cite{Survey1_TSC_Zheng}. Two primary CF methods have been followed in the literature: memory-based CF (say, memCF) and model-based CF (say, modCF). In the memCF, the similarity between users and/or services has been studied \cite{DBLP:conf/IEEEscc/LiWSZ17,DBLP:conf/icws/ZhouWGP15,sun2013personalized}. Although the memCF is simple, it has some limitations, such as low accuracy, high prediction time, low scalability \cite{Survey2_TSC}. The main reason for accuracy degradation by the memCF is its inability to handle a sparse QoS invocation log matrix. The sparsity problem is resolved by modCF, like matrix factorization \cite{lo2012extended,DBLP:conf/soca/XuYL13,DBLP:conf/wocc/LuoZXZ14,DBLP:conf/IEEEscc/QiHSGL15}, and thus, modCF achieves higher accuracy than memCF. However, the accuracy level achieved by modCF is still not up to the mark since this method barely captures the non-linear relationship between data. To improve the accuracy, hybrid methods exploiting advanced machine learning architectures \cite{TSMC_8805172,DBLP:conf/icws/XiongWLLH18} and combining the memCF and modCF \cite{DBLP:conf/icsoc/ChattopadhyayB19,chowdhury2020cahphf} have been explored further. The problem with these approaches is that they have a high prediction time \cite{Survey2_TSC}, and thereby, they are hardly acceptable in real-time systems. 

In this paper, we propose a QoS prediction framework, named FES, using multi-level clustering and multi-phase prediction algorithms. Our framework has three main modules: preprocessing, offline-learning, and online-learning. While the multi-level clustering is performed in preprocessing module, the offline and online-learning modules involve training a semi-offline regression model leveraging a 2-staged neural-network architecture. Our framework, as a whole, ensures high prediction accuracy, faster responsiveness, and high scalability.

Our earlier work on QoS prediction (CAHPHF) \cite{chowdhury2020cahphf} achieved moderately high accuracy and scalability. CAHPHF consists of two main modules: hybrid filtering and hierarchical prediction. Since both the filtering and the training of the hierarchical prediction module of CAHPHF are performed at run-time, i.e., during the prediction of the target QoS value, thus the prediction time is highly compromised. Our current work (FES) is an extension of our previous work, in which we introduce a set of significant changes to make the system robust in terms of efficiency, prediction time, and scalability. For example, unlike CAHPHF, since in FES, the target QoS value is not known apriori during preprocessing, thus instead of filtering, in this paper, we propose a multi-level clustering. The sparsity of the cluster-output is also handled during preprocessing. The preprocessing enhances the prediction accuracy without affecting the prediction time.  Additionally, the hierarchical prediction module of CAHPHF comprises two-level neural-networks (NN), where the training time of the second-level NN is considerably higher than the first-level NN. Therefore, in FES, we propose a semi-offline strategy, where the training of the second-level NN is performed in offline-mode. However, we allow the first-level NN to train in the online-mode after having the target user and service. The experimental results on the benchmark datasets show that FES achieved higher accuracy and faster responsiveness as compared to CAHPHF and other SoA methods. 

\noindent
{\textbf{Contributions}}: We now summarize our major contributions:

{{\bf{\em{(i)}}} {\emph{Attainment of Three Goals}}: 
We propose a novel framework (FES) for QoS value prediction that not only achieves higher accuracy but also is scalable and fast at the same time, in contrast to the SoA methods.

{{\bf{\em{(ii)}}} {\emph{Multi-level Clustering}}: We propose a multi-level clustering as a part of preprocessing of FES, which helps to segregate the correlated users and services to improve the performance of the framework in terms of prediction accuracy. We first perform context-aware clustering followed by similarity-based clustering, which is later merged using context-sensitive clustering. Two different modes of clustering are employed: user followed by service clustering and service followed by user clustering. Finally, the cluster-outputs are stored and used for further processing to achieve higher prediction accuracy. Since the clustering is accomplished as part of preprocessing, it does not influence the overall prediction time. Furthermore, since the large dataset is divided into smaller clusters, and the prediction algorithm needs to handle one cluster at a time at run-time, therefore FES is scalable.

{{\bf{\em{(iii)}}} {\emph{Multi-phase Prediction}}: To attain higher accuracy, FES predicts the QoS value in three different phases.
In the preprocessing module, we first employ state-of-the-art CF methods to fill-up the missing values in the QoS invocation log matrix. This is the first phase of prediction, where on the one hand, the sparsity problem is eliminated that may be encountered at the latter phase of prediction, in the absence of the first phase. On the other hand, this helps to increase the final prediction accuracy. Furthermore, since this prediction is performed in the preprocessing module of FES, the final prediction time of FES during testing is not increased for this.
We further propose a two-staged semi-offline regression model leveraging a fused neural-network architecture for the next level of prediction. While the Stage-1 of the model is responsible for QoS value prediction, the Stage-2 of the model is employed for correcting the already predicted QoS values by Stage-1. The fused neural-network architecture as a whole ensures high prediction accuracy. Furthermore, the Stage-2 of the model, which requires higher training time, is trained in the offline-mode. The training of Stage-1 is done in online-mode, which is comparatively less time-consuming. This makes our model semi-offline, which in turn ensures faster responsiveness as compared to the online framework (e.g., CAHPHF \cite{chowdhury2020cahphf}).

{{\bf{\em{(iv)}}} {\emph{Performance Analysis}}:
We analyzed the performance of FES on four benchmark datasets (WS-DREAM \cite{zheng2014investigating,DBLP:conf/issre/ZhangZL11}) to show the efficiency, scalability, and faster responsiveness of the framework with respect to the SoA methods. On average, FES achieved 19.6\% improvement on the prediction accuracy and 6.9x speed-up in prediction time as compared to the best SoA methods in the respective categories we encountered so far, while also ensuring scalability.

The rest of the paper is organized as follows. Section \ref{related} presents the brief related literature review. 
Section \ref{sec:preliminaries} discusses the problem formulation and our objectives. Section \ref{sec:method} then describes the proposed framework in detail, while the experimental analysis is presented in Section \ref{sec:result}. Finally, Section \ref{sec:conclusion} concludes our paper.

\section{Brief Literature Review}\label{related}
\noindent
QoS prediction is a well-established problem in the domain of services computing \cite{Survey2_TSC,Survey1_TSC_Zheng}. Various aspects of the QoS prediction problem have been identified, and different solutions have been proposed. 
Here, we briefly review the literature and position our work with respect to the SoA methods.

\subsection{QoS Prediction: Different Aspects}
\noindent
A reasonable amount of research focuses on QoS prediction across users 
\cite{DBLP:conf/icsoc/ZouJNWPG18, wu2017collaborative, DBLP:conf/wocc/LuoZXZ14}. 
Considering contextual information of users/services helps to improve the accuracy of QoS prediction, a location-aware QoS prediction problem has been studied further \cite{DBLP:conf/icws/LeePB15,DBLP:conf/compsac/ChenMSXL17,DBLP:journals/soca/YuH16,tang2012location}. 
Sometimes, the location information of users/services has been used as contextual information to cluster the users and/or services \cite{chowdhury2020cahphf}. Often the location information has been exploited as part of the feature vector to predict the target QoS value \cite{TSMC_8805172}. Some research articles also explore time-aware QoS prediction \cite{DBLP:journals/soca/YuH16,DBLP:conf/icws/WuQWZY16,DBLP:conf/issre/ZhangZL11,DBLP:journals/computing/KeshavarziHB20,DBLP:journals/access/LiWLR18,6928878,DBLP:journals/tsc/FanHZWBC21}. The multi-dimensional QoS prediction problem has been studied as well \cite{DBLP:journals/tsc/WangMCYC19}.
In this paper, we focus on location-aware QoS prediction.


\subsection{QoS Prediction: Various Solutions}
\noindent
In this sub-section, we discuss the strengths and weaknesses of different methods to address various issues in QoS prediction.

\medskip
\noindent
{\textbf{Memory-based CF (Similarity-based Methods)}}:
One of the most prevalent methods for QoS prediction is collaborative filtering. 
The memory-based CF (memCF) \cite{DBLP:conf/IEEEscc/LiWSZ17,DBLP:conf/icws/ZhouWGP15,sun2013personalized} is a preliminary version of CF, where the pairwise similarity between users \cite{DBLP:journals/tsc/ZhengMLK13} and/or services \cite{sarwar2001item} is measured to predict the target QoS value \cite{DBLP:conf/icws/WuQWZY16,wu2017collaborative}. Pearson-correlation coefficient \cite{DBLP:journals/tsc/ZhengMLK13}, cosine similarity \cite{DBLP:conf/icsoc/ChattopadhyayB19}, ratio-based similarity \cite{DBLP:conf/icsoc/ZouJNWPG18} are some popular metrics to compute the similarity. Although the memCF is easy to implement, it hardly achieves plausible prediction accuracy due to the sparsity problem in the QoS invocation log matrix. Moreover, it suffers from a scalability issue since a considerable number of similarity computations are required to be performed for a dataset with a large number of users and services \cite{Survey2_TSC}. Furthermore, cold-start problem \cite{Survey2_TSC} is another major limitation of memCF. 

%
\medskip
\noindent
{\textbf{Model-based CF (Clustering-based Method)}}:
To improve the scalability, as an improvisation of memCF, clustering-based methods have been introduced. The idea here is to perform clustering on the set of users and/or services before QoS prediction \cite{DBLP:conf/icws/ShiZLC11,DBLP:journals/soca/YuH16}. As clustering is done in offline mode, the scalability issue is partially resolved. However, the cold-start problem, low-accuracy still persist for this category \cite{Survey2_TSC}.


%
\medskip
\noindent
{\textbf{Model-based CF (Matrix Factorization)}}:
To increase prediction accuracy and to deal with the sparsity problem and the cold-start problem, matrix factorization (MF) has been adopted \cite{DBLP:journals/tsc/ZhengMLK13}.
MF decomposes the QoS invocation log matrix into low-rank matrices and then reconstructs the same. 
Empirically, the MF-based methods are better than the 
early-mentioned 
approaches in terms of prediction accuracy and scalability.
As an enhancement of traditional MF, regularization has been incorporated further \cite{lo2012extended,DBLP:conf/soca/XuYL13,DBLP:conf/wocc/LuoZXZ14,DBLP:conf/IEEEscc/QiHSGL15}. However, the prediction accuracy is yet to achieve a satisfactory level, since MF-based methods are unable to capture the higher-order, non-linear relationship between users and services in terms of QoS values \cite{9189762}. 
Due to the online training, in many cases, MF-based methods experience higher prediction time, and thereby, barely suit for a real-time system.

\medskip
\noindent
{\textbf{Model-based CF (Factorization Machine)}}:
Factorization Machine (FM) usually casts the QoS prediction to a regression problem. In general, FM performs better than MF in terms of prediction accuracy, since most of the time, it can capture the non-linearity between QoS values of services invoked by different users \cite{DBLP:conf/icsoc/WuXCCZ17,access/LiLZ20b,9189762}.  

Some deep learning-based methods also have been explored for QoS prediction in the literature. For example, in \cite{TSMC_8805172}, multi-layer perceptron (MLP) was employed. The identifier and location information of users and services were embedded as a one-hot encoding vector, which was fed to the neural network as input.
To address the sparsity problem, an encoder-decoder architecture was used in \cite{Enc_Dec_ECSOC}. 
A stacked auto-encoder was explored in \cite{AEpercom2019} to enrich the feature representation, and thus, improving the prediction accuracy. 
Yin et al. \cite{DBLP:journals/access/YinZXZMY19} proposed an ensemble method, where an auto-encoder architecture was used
to alleviate the cold start problem. 
To capture the non-linear relationship between the features, a neural network model with radial basis function was employed in \cite{DBLP:conf/IEEEscc/ZhangSL0L16}. 
Xiong et al. \cite{DBLP:conf/icws/XiongWLLH18} adopted long short-term memory (LSTM) to improve the efficiency of the prediction model.
%
Although the above deep learning-based models aim to solve a specific problem, often ignore the others. For example, the correlation between the users and services is ignored mostly by these methods. Therefore, the prediction accuracy hardly meets the expectation.

\medskip
\noindent
{\textbf{Hybrid Methods}}:
Hybrid methods combine both the memCF and modCF to achieve high accuracy \cite{DBLP:conf/icsoc/ChattopadhyayB19,chowdhury2020cahphf}. Evidently, these methods are highly efficient. However, the frameworks of \cite{chowdhury2020cahphf} and \cite{DBLP:conf/icsoc/ChattopadhyayB19} require online training. Thus, these methods take significant time for prediction, which in turn makes the methods inefficient for a real-time system.

\medskip
\noindent
{\textbf{Trade-off between Efficiency, Responsiveness, Scalability}}:
In QoS prediction, we clearly observe the trade-off between accuracy, responsiveness, and scalability. The methods \cite{DBLP:conf/icsoc/ChattopadhyayB19,chowdhury2020cahphf}, which achieved higher accuracy, compromised significantly on responsiveness/scalability. On the other hand, the methods \cite{wu2017collaborative,DBLP:journals/soca/YuH16,DBLP:conf/wocc/LuoZXZ14}, which are faster enough to be part of a real-time system, compromised prediction accuracy. In this paper, we study this trade-off and propose a framework to balance efficiency, scalability, and responsiveness.

\subsection{Positioning Our Work (FES)}
\noindent
In contrast to the SoA methods, our framework (FES) attempts to address the above issues. 
We now discuss the different goals of QoS prediction and the strategies undertaken in FES to achieve these.
\begin{enumerate}[labelsep = *,leftmargin=*]
 \item[{\textbf{{\em(i)}}}] {\textbf{Efficiency}}: To increase the prediction accuracy of FES, we use a multi-phase prediction strategy. Additionally, we address the following issues, which 
 impedes the system accuracy mostly.
 \begin{itemize}
  \item {\emph{Handling Sparsity Problem}}: This problem is resolved in preprocessing module, where we use 
  off-the-shelf 
  CF and MF methods to fill-up the missing values in the QoS invocation log matrix, which is further used in the next-level of our framework. 
  \item {\emph{Capturing correlation}}: We exploit multi-level clustering to partition the correlated users and services.
  \item {\emph{Capturing non-linearity among dataset}}: We model a non-linear regression framework by leveraging a two-staged neural network architecture.
 \end{itemize} 
 \item[{\textbf{{\em(ii)}}}] {\textbf{Fast Responsiveness}}: To ensure the faster prediction time of our framework, we propose a semi-offline training architecture and thereby, make FES suitable for a real-time recommendation system.
 \item[{\textbf{{\em(iii)}}}] {\textbf{Scalability}}: We ensure the scalability of our framework by dividing the large dataset into multiple clusters.
 \item[{\textbf{{\em(iv)}}}] {\textbf{Cold-start problem}}: We address the cold-start problem by considering the contextual (specifically, location) information of users and services while clustering them. 
\end{enumerate}

\noindent
Although FES has many advantages over the SoA methods, it has a few limitations as well. For example, FES does not deal with the time-aware QoS prediction or other aspects for QoS prediction (e.g., the variation of QoS across distributed execution platforms), which we aim to address in the future.

\section{Overview and Problem Formulation}\label{sec:preliminaries}
\noindent
In this section, we formalize our problem statement. We are given the following input parameters:

\begin{itemize}
     \item A set of users ${\cal{U}} = \{u_1, u_2, \ldots, u_n\}$
     \item A set of services ${\cal{S}} = \{s_1, s_2, \ldots, s_m\}$
     \item A QoS parameter $q$
     \item A QoS invocation log ${\cal{Q}}$ in the form of a 2D matrix
     \item A contextual information $\alpha_i$ for each $u_i \in {\cal{U}}$
     \item A contextual information $\beta_j$ for each $s_j \in {\cal{S}}$ 
     \item A set of services ${\cal{S}}_i \subseteq {\cal{S}}$ invoked by each $u_i \in {\cal{U}}$
     \item A set of users ${\cal{U}}_j \subseteq {\cal{U}}$ that invoked each $s_j \in {\cal{S}}$
     \item A query ${\cal{R}} = (u_{t_1}, s_{t_2})$ in the form of a target user and a target service 
\end{itemize}
\noindent
Before describing our objective, we first discuss the details of a few input parameters.
We begin with the description of the QoS invocation log matrix. Each row and column of ${\cal{Q}}$ correspond to a user and a service, respectively. Each entry of ${\cal{Q}}$, i.e., ${\cal{Q}}[i, j]$ represents the value of $q$ of $s_j$ invoked by $u_i$. If a user $u_i$ accessed a service $s_j$ in the past, the corresponding value of $q$ is instantaneously recorded in ${\cal{Q}}$. However, each entry of ${\cal{Q}}$ is not a valid entry. An invalid entry of ${\cal{Q}}$ is represented by 0, which implies the corresponding user has never invoked the respective service. It may be noted that ${\cal{Q}}$ is mostly a sparse matrix.  

Furthermore, in this paper, the contextual information of a user/service refers to the location information only. The location, here, is represented by a 2-tuple $(\phi_i, \psi_i)$, where $\phi_i$ and $\psi_i$ represent the latitude and longitude of the user/service, respectively.

{\textbf{Objective}}: Given a query ${\cal{R}}$, the objective of the classical QoS prediction problem is to predict the value of $q$ of $s_{t_2}$ to be invoked by $u_{t_1}$ (often, termed as the {\emph{target QoS}}). However, in this paper, our objective is to design a robust framework that not only can predict the QoS value accurately but also in a reasonable time limit to be adopted in a real-time service recommendation system. We now formally define three important objectives of our framework:

\begin{itemize}
 \item {\emph{Goal-1 (Efficiency)}}: The prediction algorithm should be efficient in terms of prediction accuracy.
 \item {\emph{Goal-2 (Responsiveness)}}: The algorithm has to be faster (in terms of prediction time) enough to be incorporated into a real-time system.
 \item {\emph{Goal-3 (Scalability)}}: The algorithm needs to be scalable so that it can handle a significantly large dataset in a reasonable time-limit.
\end{itemize}

\noindent 
In the next section, we discuss our solution framework.


\section{Detailed Methodology}\label{sec:method}
\noindent
In this paper, we develop a QoS prediction framework (FES) based on a given QoS invocation log matrix ${\cal{Q}}$. The main idea of this work is to preprocess ${\cal{Q}}$ at first, followed by creating a semi-offline model to predict the target QoS value in real-time. FES comprises three modules: preprocessing, offline-learning, and online-learning, as shown in  
Fig. \ref{fig:flow}. We now illustrate these modules in the subsequent subsections.

\begin{figure}[!t]
    \centering
 	\includegraphics[width=\linewidth]{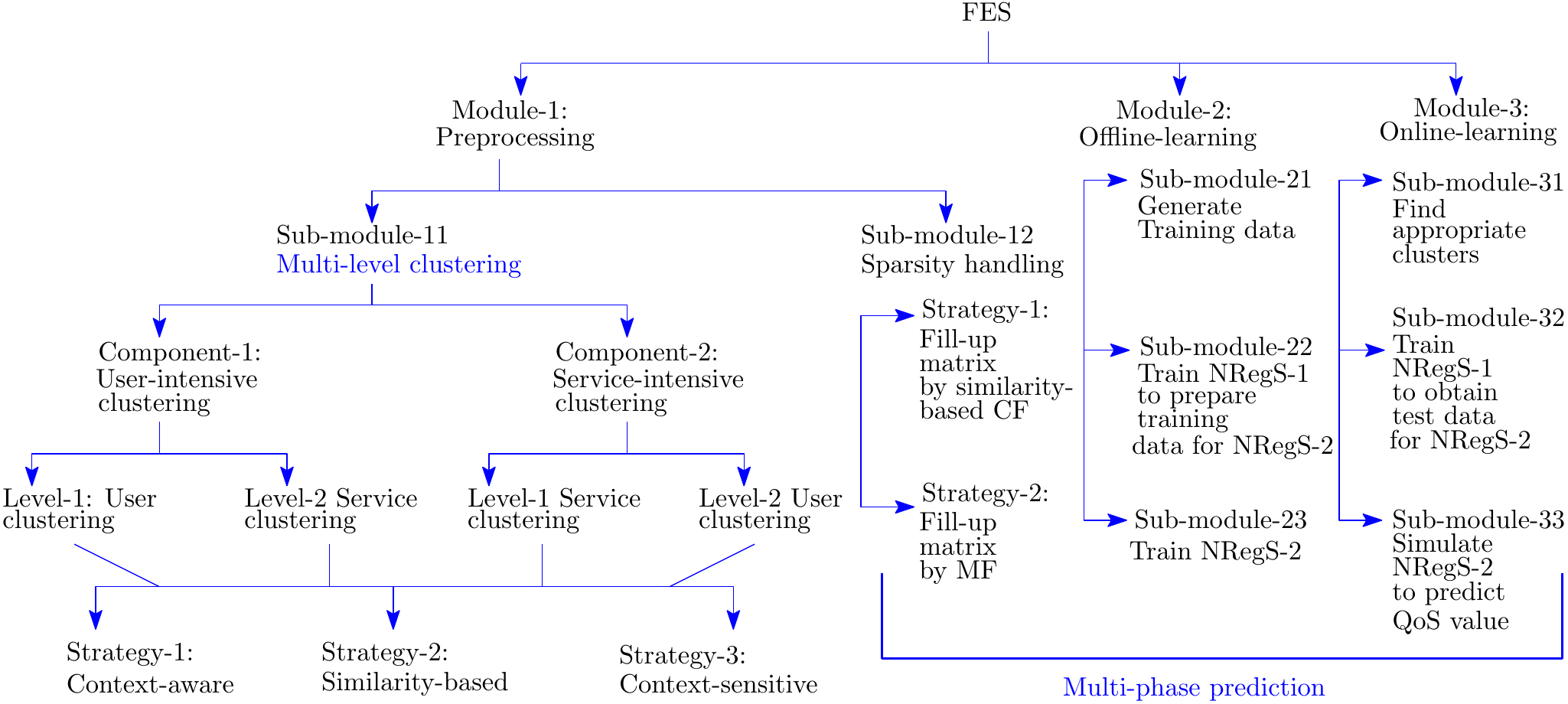}
 	\caption{Different modules of FES}
 	\label{fig:flow}
\end{figure}
\begin{figure*}
    \centering
 	\includegraphics[width=0.7\linewidth]{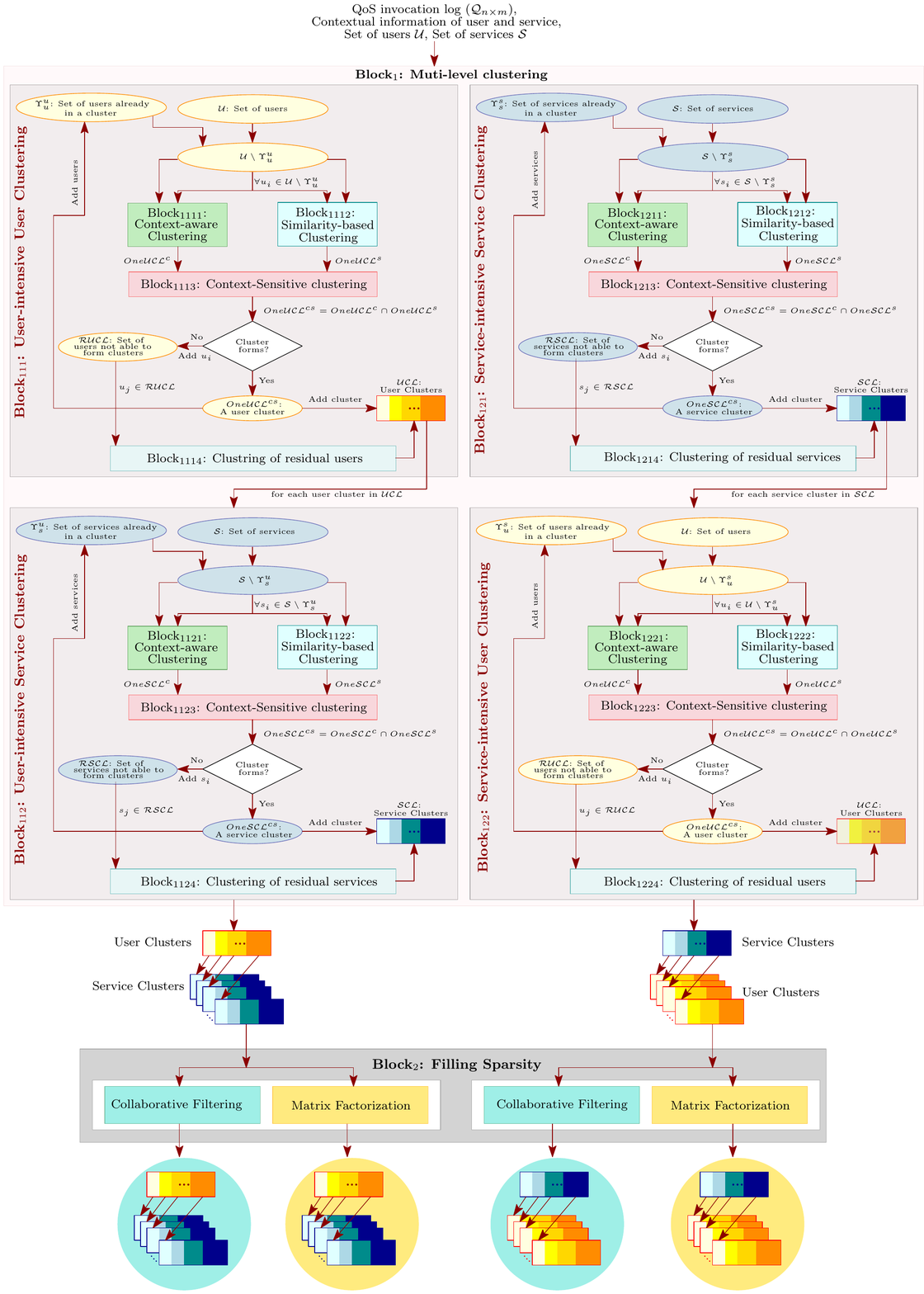}
 	\caption{Preprocessing module of FES}
 	\label{fig:architecture}
\end{figure*}

\subsection{Preprocessing Module}
\noindent
The objective of this module is to process ${\cal{Q}}$ so that it can be utilized further in the next-part of FES to improve the prediction accuracy. Fig. \ref{fig:architecture} shows the pictorial overview of the preprocessing of FES. The preprocessing module has two sub-modules: multi-level clustering and sparsity handling, which are discussed now.

\subsubsection{Multi-level Clustering}\label{subsubsec:mlc}
\noindent
The first sub-module of the preprocessing includes the formation of multi-level clusters from the set of users and services (refers to Block$_1$ of Fig. \ref{fig:architecture}), which involves two components: (a) user-intensive multi-level clustering (say, UICL), and (b) service-intensive multi-level clustering (say, SICL). It may be noted that the multi-level clusters generated by UICL (say, ${\cal{CL}}_{UI}$: having user cluster followed by the service cluster) are different from the one obtained from SICL (say, ${\cal{CL}}_{SI}$: having service cluster followed by the user cluster). We begin with discussing the details of UICL.

a) {\emph{User-intensive Multi-level Clustering (UICL):}} 
Algorithms \ref{algo:uiucl} and \ref{algo:uiscl} present the formal algorithm for UICL. 
Since UICL involves user clustering (Refers to Block$_{111}$ of Fig. \ref{fig:architecture}), followed by service clustering (Refers to Block$_{112}$ of Fig. \ref{fig:architecture}), it is referred to as multi-level clustering. 
Three different strategies are combined to form user/service clustering. 
We now discuss each of these three strategies below.

\begin{algorithm}\scriptsize
  \caption{$UI\_UserClustering$}
  \begin{algorithmic}[1]
    \State {\bf{Input:}} A set of users $\cal{U}$, A set of services ${\cal{S}}$
    \State {\bf{Output:}} A set of clusters ${\cal{CL}}_{UI}$
    
    \State {\bf{Initialization:}} ${\cal{CL}}_{UI} = NULL$, ${\cal{UCL}} = 
NULL$, ${\cal{RUCL}} = NULL$, $\Upsilon^u_u = NULL$, $clusterCount = 0$;

    \Comment{$\Upsilon^u_u$ represents the set of users that are already in some clusters}\\
    \Comment{${\cal{RUCL}}$ represents the set of users that cannot form a cluster}
    \For {each $u_i \in {\cal{U}} \setminus (\Upsilon^u_u \cup {\cal{RUCL}})$}
        \State $One{\cal{UCL}}^c \leftarrow \{u_i\}$;\Comment{For context-aware clustering}
        \State $One{\cal{UCL}}^s \leftarrow \{u_i\}$;\Comment{For similarity-based clustering}
        \State $\Upsilon_1 = NULL$;
        \If {$\left(\left|{\cal{U}}\right| - \left|\Upsilon^u_u \cup 
{\cal{RUCL}}\right|\right) \le T^u_{cs}$}
            \State break;
        \EndIf
        \For {each $u_j \in One{\cal{UCL}}^c \setminus \Upsilon_1$}\Comment{Context-aware clustering}
            \State $\Upsilon_1 \leftarrow \Upsilon_1 \cup \{u_j\}$;
            \For {each $u_k \in ({\cal{U}} \setminus \Upsilon^u_u) \setminus 
One{\cal{UCL}}^c$} 
                \If {$HD (\alpha_j, \alpha_k) \le T^u_c$}
                    \State $One{\cal{UCL}}^c \leftarrow One{\cal{UCL}}^c \cup 
\{u_k\}$;
                \EndIf
            \EndFor
        \EndFor
        \State $\Upsilon_1 = NULL$;
        \For {each $u_j \in One{\cal{UCL}}^s \setminus \Upsilon_1$}\Comment{Similarity-based clustering}
            \State $\Upsilon_1 \leftarrow \Upsilon_1 \cup \{u_j\}$;
            \For {each $u_k \in ({\cal{U}} \setminus \Upsilon^u_u) \setminus 
One{\cal{UCL}}^s$}
                \If {$CSM (u_j, u_k) \ge T^u_s$}
                    \State $One{\cal{UCL}}^s \leftarrow One{\cal{UCL}}^s \cup 
\{u_k\}$;
                \EndIf
            \EndFor
        \EndFor
        \If {$One{\cal{UCL}}^c \cap One{\cal{UCL}}^s \ge T^u_{cs}$}\Comment{Context-sensitive clustering}
            \State ${\cal{UCL}}[clusterCount] = One{\cal{UCL}}^c \cap 
One{\cal{UCL}}^s$;
            \State $clusterCount~+=~1$;
            \State $\Upsilon^u_u \leftarrow \Upsilon^u_u \cup 
\left(One{\cal{UCL}}^c \cap One{\cal{UCL}}^s\right)$;
        \Else
            \State ${\cal{RUCL}} \leftarrow {\cal{RUCL}} \cup \{u_i\}$;
        \EndIf        
    \EndFor
    \For {each $u_i \in {\cal{RUCL}} \cup ({\cal{U}} \setminus \Upsilon^u_u)$}\Comment{Clustering of residual users}
        \State $maxSimilarity = 0$, $cluster = 0$;
        \For {$k = 0$ to $clusterCount$}
            \State $maxClusterSimilarity \leftarrow \max_{u_j \in 
{\cal{UCL}}[k]} CSM (u_i, u_j)$;
            \If {$maxSimilarity < maxClusterSimilarity$}
                \State $maxSimilarity = maxClusterSimilarity$; $cluster = k$;
            \EndIf
        \EndFor
        \State ${\cal{UCL}}[cluster] \leftarrow {\cal{UCL}}[cluster] \cup 
\{u_i\}$;
    \EndFor
    
    \For {$k = 0$ to $clusterCount$}\Comment{Creating the second-level cluster}
        \State ${\cal{CL}}_{UI}[k] \leftarrow \left({\cal{UCL}}[k], 
UI\_ServiceClustering ({\cal{UCL}}[k], {\cal{S}})\right)$; \\
        \Comment{UI\_ServiceClustering refers to 
user-intensive service clustering algorithm}
    \EndFor    
    \State return ${\cal{CL}}_{UI}$;
  \end{algorithmic}
  \label{algo:uiucl}
\end{algorithm}

\begin{algorithm}\scriptsize
  \caption{$UI\_ServiceClustering$}
  \begin{algorithmic}[1]
    \State {\bf{Input:}} A set of users ${\cal{U}}' \subseteq {\cal{U}}$, A set 
of services ${\cal{S}}$
    \State {\bf{Output:}} A set of clusters ${\cal{SCL}}$
    
    \State {\bf{Initialization:}} ${\cal{SCL}} = NULL$, ${\cal{RSCL}} = NULL$, 
$\Upsilon^u_s = NULL$, $clusterCount = 0$;
    
    \For {each $s_i \in {\cal{S}} \setminus (\Upsilon^u_s \cup {\cal{RSCL}})$}
        \State $One{\cal{SCL}}^c \leftarrow \{s_i\}$; \Comment{For context-aware clustering}
        \State $One{\cal{SCL}}^s \leftarrow \{s_i\}$; \Comment{For similarity-based clustering}
        \State $\Upsilon_1 = NULL$;
        \If {$\left(\left|{\cal{S}}\right| - \left|\Upsilon^u_s \cup 
{\cal{RSCL}}\right|\right) \le T^s_{cs}$}
            \State break;
        \EndIf
        \For {each $s_j \in One{\cal{SCL}}^c \setminus \Upsilon_1$} \Comment{Context-aware clustering}
            \State $\Upsilon_1 \leftarrow \Upsilon_1 \cup \{s_j\}$;
            \For {each $s_k \in ({\cal{S}} \setminus \Upsilon^u_s) \setminus 
One{\cal{SCL}}^c$} 
                \If {$HD (\beta_j, \beta_k) \le T^s_c$}
                    \State $One{\cal{SCL}}^c \leftarrow One{\cal{SCL}}^c \cup 
\{s_k\}$;
                \EndIf
            \EndFor
        \EndFor
        \State $\Upsilon_1 = NULL$;
        \For {each $s_j \in One{\cal{SCL}}^s \setminus \Upsilon_1$}\Comment{Similarity-based clustering}
            \State $\Upsilon_1 \leftarrow \Upsilon_1 \cup \{s_j\}$;
            \For {each $s_k \in ({\cal{S}} \setminus \Upsilon^u_s) \setminus 
One{\cal{SCL}}^s$}
                \If {$CSM (s_j, s_k) \ge T^s_s$}
                    \State $One{\cal{SCL}}^s \leftarrow One{\cal{SCL}}^s \cup 
\{s_k\}$;
                \EndIf
            \EndFor
        \EndFor
        \If {$One{\cal{SCL}}^c \cap One{\cal{SCL}}^s \ge T^s_{cs}$}\Comment{Context-sensitive clustering}
            \State ${\cal{SCL}}[clusterCount] = One{\cal{SCL}}^c \cap 
One{\cal{SCL}}^s$;
            \State $clusterCount~+=~1$;
            \State $\Upsilon^u_s \leftarrow \Upsilon^u_s \cup 
\left(One{\cal{SCL}}^c \cap One{\cal{SCL}}^s\right)$;
        \Else
            \State ${\cal{RSCL}} \leftarrow {\cal{RSCL}} \cup \{s_i\}$;
        \EndIf        
    \EndFor
    \For {each $s_i \in {\cal{RSCL}} \cup ({\cal{S}} \setminus \Upsilon^u_s)$}\Comment{Clustering of residual services}
        \State $maxSimilarity = 0$, $cluster = 0$;
        \For {$k = 0$ to $clusterCount$}
            \State $maxClusterSimilarity \leftarrow \max_{s_j \in 
{\cal{SCL}}[k]} CSM (s_i, s_j)$;
            \If {$maxSimilarity < maxClusterSimilarity$}
                \State $maxSimilarity = maxClusterSimilarity$; $cluster = k$;
            \EndIf
        \EndFor
        \State ${\cal{SCL}}[cluster] \leftarrow {\cal{SCL}}[cluster] \cup 
\{s_i\}$;
    \EndFor  
    \State return ${\cal{SCL}}$;
  \end{algorithmic}
  \label{algo:uiscl}
\end{algorithm}

i) {\emph{Context-aware Clustering:}} 
In each iteration of the user clustering of UICL, a new user $u_i \in {\cal{U}}$ is selected to form a cluster. At first, $u_i$ is used to finding a cluster (say, $One{\cal{UCL}}^c$) containing a set of users contextually similar to $u_i$. Here, we use the Haversine distance (HD) function \cite{chowdhury2020cahphf} to find out contextually similar users. $One{\cal{UCL}}^c$ is generated iteratively. In each iteration, we obtain a set of users that are contextually similar to at-least one user in the cluster formulated so far and 
include them in the cluster at the end of the iteration. Initially, the cluster contains only $u_i$. The clustering algorithm terminates once no new user is found to be added. Therefore, for each user $u_j \in One{\cal{UCL}}^c$, there exists at least one user $u_k \in One{\cal{UCL}}^c$, such that the Haversine distance between the contextual information of $u_j$ and $u_k$ (i.e., $HD(\alpha_j, \alpha_k)$)  is less than a given threshold $T^u_c$. Lines 12 to 19 of Algorithm \ref{algo:uiucl} present the formal description of context-aware clustering (refers to Block$_{1111}$ of Fig. \ref{fig:architecture}).
 
ii) {\emph{Similarity-based Clustering:}}
$u_i$ is then employed to find a cluster (say, $One{\cal{UCL}}^s$) containing users behaviorally
similar to $u_i$. Cosine similarity measure (CSM) \cite{chowdhury2020cahphf} is used to find out the behavioral similarity between two users in terms of their QoS invocation logs. This strategy is similar to the previous one. A given parameter $T^u_s$ is used here to decide the similarity threshold. Lines 21 to 28 of Algorithm \ref{algo:uiucl} formally describe similarity-based clustering (refers to Block$_{1112}$ of Fig. \ref{fig:architecture}).
 
iii) {\emph{Context-sensitive Clustering:}} 
In this step, the intersection between $One{\cal{UCL}}^c$ and $One{\cal{UCL}}^s$ is computed. If the size of the intersection set is more than a given threshold $T^u_{cs}$, the cluster is considered and added to ${\cal{UCL}}$. Otherwise, $u_i$ is added to the set of residual users (say, ${\cal{RUCL}}$). Lines 29 to 35 of Algorithm \ref{algo:uiucl} present the description of context-sensitive clustering (refers to Block$_{1113}$ of Fig. \ref{fig:architecture}).

iv) {\emph{Clustering of Residual Users:}} 
Once the user cluster generation is done, the remaining users, which do not belong to any cluster, are dealt with to place in the appropriate clusters. For each such user $u_i$, we compute the maximum cosine similarity measure between $u_i$ and every other user in one cluster. Finally, we add $u_i$ to the cluster having maximum similarity value computed across all clusters.  
Lines 37 to 46 of Algorithm \ref{algo:uiucl} formally present this step (refers to Block$_{1114}$ of Fig. \ref{fig:architecture}).
Likewise, if the dataset contains a new user $u_i$ without any QoS record, we compute the minimum Haversine distance between $u_i$ and every other user in one cluster. Finally, we add $u_i$ to the cluster having minimum distance across all clusters. This step completes the user clustering of UICL. We now discuss the service clustering of UICL.

v) {\emph{Second-level Clustering:}}
Once user clusters are generated, for each user cluster ${\cal{UCL}}[k] \in {\cal{UCL}}$, a set of service clusters (say, ${\cal{SCL}}$) are generated further. Algorithm \ref{algo:uiscl} formally presents the user-intensive service clustering. The service clustering algorithm is similar to the user clustering algorithm. The only difference is that in the similarity-based clustering strategy of service clustering, the QoS invocation log corresponding to the users in ${\cal{UCL}}[k]$ is used to compute the cosine similarity measure. Finally, ${\cal{UCL}}[k] \text{ and } {\cal{SCL}}$ are added to ${\cal{CL}}_{UI}$. Therefore, it may be noted that the  clusters obtained from UICL (i.e., ${\cal{CL}}_{UI}$) have two-levels, where the first level contains the user clusters, and corresponding to each user cluster, the second level comprises the service clusters.

It may be noted that each user-intensive multi-level cluster (i.e., a user cluster followed by a service cluster) corresponds to a unique QoS invocation log matrix. In other words, the multi-level clusters are mutually exclusive in terms of user-service pair. Further, if we combine the QoS invocation logs corresponding to each user-intensive multi-level cluster, we are able to generate the given ${\cal{Q}}$. In other words, the multi-level clusters are collectively exhaustive.

b) {\emph{Service-intensive Multi-level Clustering (SICL):}} 
SICL is algorithm-wise similar to UICL. The only difference here is that the service clustering is performed before the user clustering. Unlike UICL, the clusters obtained from SICL (i.e., ${\cal{CL}}_{SI}$) comprises the service clusters in the first level and user clusters in the second level corresponding to each service cluster. However, similar to UICL, the service-intensive multi-level clusters are also mutually exclusive and collectively exhaustive.

\subsubsection{Sparsity Handling}
\noindent
The QoS invocation logs are, in general, sparse in nature, which means most of the entries of the QoS invocation logs are invalid. The prediction accuracy significantly degrades due to the sparsity. The objective of this module is to fill up the sparse matrices obtained from the multi-level clustering to attain higher accuracy.  

Two off-the-shelf QoS prediction methods are adopted here to fill up the sparse matrices obtained from the previous sub-module of preprocessing, namely similarity-based collaborative filtering (CF) and matrix factorization (MF) \cite{chowdhury2020cahphf}. Therefore, it may be noted that for each QoS invocation log matrix corresponding to a multi-level cluster in ${\cal{CL}}_{UI}$ and ${\cal{CL}}_{SI}$, two copies of invocation logs are stored, one is filled up by CF \cite{chowdhury2020cahphf}, and the other is filled by MF \cite{chowdhury2020cahphf}. Once this step is done, for each user-service pair $(u_i, s_j)$, four QoS log matrices continue to exist containing the pair, as shown in Table \ref{tab:rp}.

%

\begin{table}[!h]
 \centering
 \scriptsize
 \caption{QoS log matrices generated in preprocessing}
 \begin{tabular}{c|c|c|c|c}
 \hlineB{2.5}
 QoS matrix & User set & Service set & Generated by & Filled-up by\\
 \hlineB{2.5}
 ${\cal{Q}}_{u}^{c}$ & ${\cal{U}}_{u}^{c}$ ($\ni u_i$) & ${\cal{S}}_{u}^{c}$ ($\ni s_j$) & UICL & CF\\
 \hline
 ${\cal{Q}}_{u}^{m}$ & ${\cal{U}}_{u}^{m}$ ($\ni u_i$) & ${\cal{S}}_{u}^{m}$ ($\ni s_j$) & UICL & MF\\
 \hline
 ${\cal{Q}}_{s}^{c}$ & ${\cal{U}}_{s}^{c}$ ($\ni u_i$) & ${\cal{S}}_{s}^{c}$ ($\ni s_j$) & SICL & CF\\
 \hline
 ${\cal{Q}}_{s}^{m}$ & ${\cal{U}}_{s}^{m}$ ($\ni u_i$) & ${\cal{S}}_{s}^{m}$ ($\ni s_j$) & SICL & MF\\
 \hlineB{2.5}
 \multicolumn{5}{r}{${\cal{U}}_{u}^{c}$ ($\ni u_i$) means $u_i \in {\cal{U}}_{u}^{c}$}
 \end{tabular}
 \label{tab:rp}
\end{table}

\begin{figure}
    \centering
 	\includegraphics[width=\linewidth]{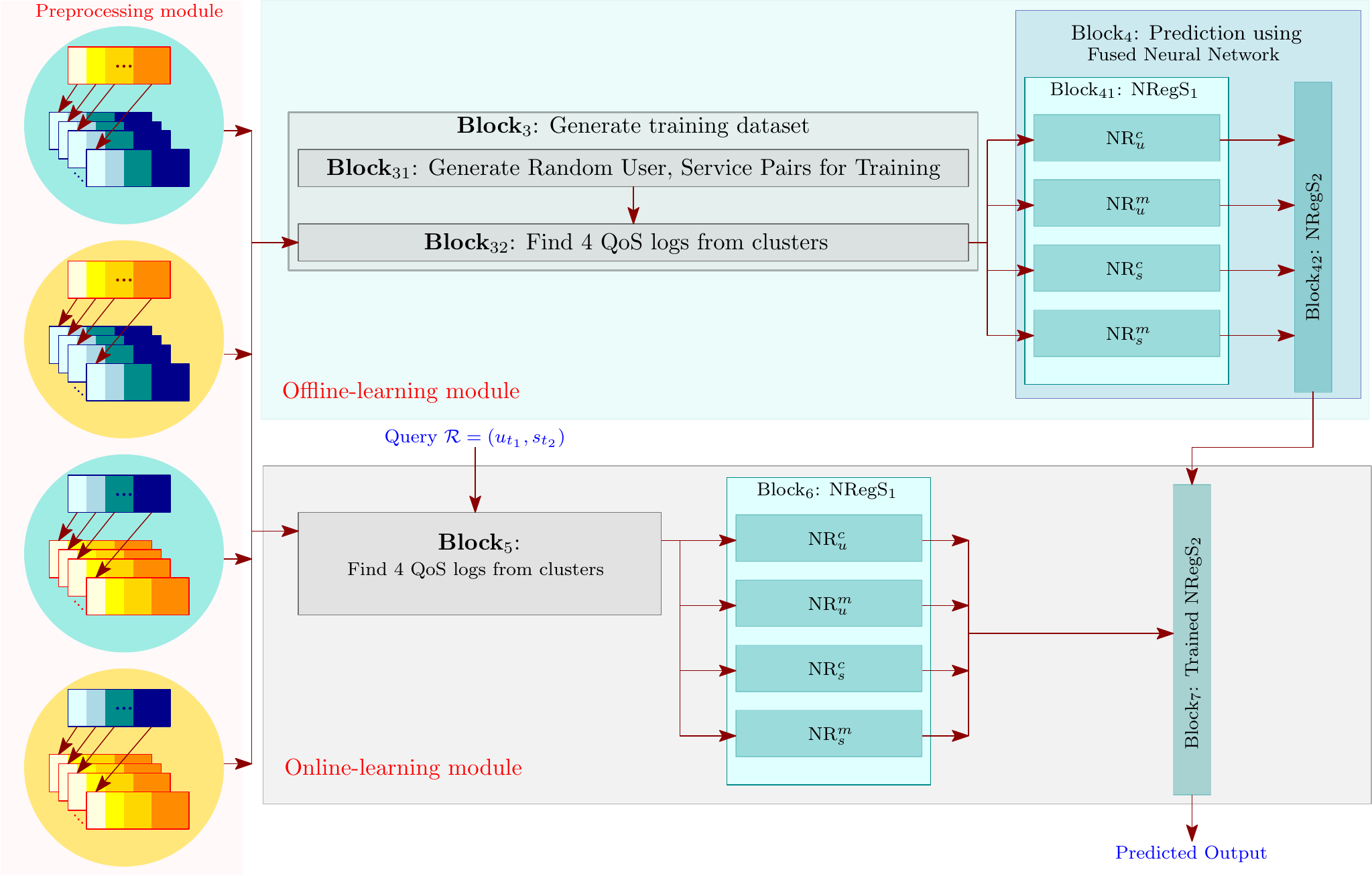}
 	\caption{Offline and online-learning modules of FES}
 	\label{fig:architecture2}
\end{figure}

\subsection{Offline-learning Module}
\noindent
We employ a fused neural-network architecture (say, fuNN) for the final QoS prediction. The fuNN has two stages: Neural-Regression Stage-1 (NRegS$_1$) and Neural-Regression Stage-2 (NRegS$_2$). While NRegS$_1$ aims to predict the target QoS value, the objective of NRegS$_2$ is to improve the prediction accuracy. NRegS$_1$ consists of four neural-networks: $NR_{u}^{c}$, $NR_{u}^{m}$, $NR_{s}^{c}$, and $NR_{s}^{m}$. The purpose of each $NR_x^y$ is to predict the target QoS value from the preprocessed matrix ${\cal{Q}}_x^y$, where $x \in \{u, s\}$ and $y \in \{c, m\}$. Once the four values are predicted for the target QoS, the aim of NRegS$_2$ is to aggregate the predicted values obtained from the four neural-networks to improve the prediction accuracy. Fig. \ref{fig:architecture2} shows the pictorial overview of the offline-learning module. Before discussing the details of fuNN, we first discuss the details of NRegS$_1$ and NRegS$_2$.

{\textbf{Details of NRegS$_1$}}: NRegS$_1$ is an online-training model. Given a user-service pair $(u_i, s_j)$, the objective of NRegS$_1$ is to predict four different QoS values of $s_j$ to be invoked by $u_i$ from four different QoS log matrices. 
Each target user-service pair creates four instances of neural networks $NR_x^y$ (for $x \in \{u, s\}$ and $y \in \{c, m\}$), each of which exploits their respective ${\cal{Q}}_x^y$ to predict the corresponding value. We now discuss the details of $NR_x^y$. 
Each $NR_x^y$ is a fully connected feed-forward neural-network with back-propagation \cite{lathuiliere2019comprehensive}. Each training sample of $NR_x^y$ consists of the service invocation pattern of user $u_k (\ne u_i) \in {\cal{U}}_x^y$. The service-invocation pattern of $u_k$ consists of the QoS values of the services in ${\cal{S}}_x^y$ other than $s_j$. The details of the training-testing dataset of $NR_x^y$ is formally specified below:

\noindent
{\emph{Training Data}}: $\forall u_k \in {\cal{U}}_x^y \setminus \{u_i\}$, $\{({\cal{Q}}_x^y[k, l]) ~|~ \forall s_l \in {\cal{S}}_x^y \setminus \{s_j\}\}$\\
{\emph{Target Value}}: $\forall u_k (\ne u_i) \in {\cal{U}}_x^y$, ${\cal{Q}}_x^y[k, j]$\\
{\emph{Test Vector}}: for given $u_i$, $\{({\cal{Q}}_x^y[i, l]) ~|~ \forall s_l \in {\cal{S}}_x^y \setminus \{s_j\}\}$\\
{\emph{Value to be Predicted}}: ${\cal{Q}}_x^y[i, j]$

Therefore, the training dataset of $NR_x^y$ contains $|{\cal{U}}_x^y| - 1$ number of training samples, while each sample consists of $|{\cal{S}}_x^y| - 1$ number of features.

{\textbf{Details of NRegS$_2$}}: NRegS$_2$ is an offline-training model comprising a single fully-connected feed-forward neural network with back-propagation. 
The primary goal of NRegS$_2$ is to increase the prediction accuracy by combining the four predicted values obtained from NRegS$_1$. Each training sample of NRegS$_2$, therefore, is represented by four features.
The actual QoS value of the given user-service pair is the target. 


{\textbf{Details of fuNN}}: The fuNN is a semi-offline training model combining NRegS$_1$ and NRegS$_2$. The NRegS$_1$ works on online-mode after having the information of the target user-service pair. NRegS$_2$, on the other hand, aims to combine the outputs produced by NRegS$_1$. Therefore, to generate each training sample in the training dataset for NRegS$_2$, all four neural-networks of NRegS$_1$ need to be trained. Therefore, generating the training dataset for NRegS$_2$ together with training the NRegS$_2$ is significantly time-consuming.
The idea of fuNN is to train NRegS$_2$ in offline-mode, which makes fuNN a semi-offline model. The architectural configurations of 
NRegS$_1$ and NRegS$_2$ are given in Section \ref{subsec:config}.

{\textbf{Work-flow of Offline-learning Module}}:
We now explain the functioning of sub-modules of the offline-learning module.

{\textbf{\em{(i)}}} The task of the first sub-module (refers to Block$_{31}$ of Fig. \ref{fig:architecture2}) is to collect user-service pairs uniformly distributed across all multi-level clusters in ${\cal{CL}}_{UI}$ and ${\cal{CL}}_{SI}$ having actual value recorded in the QoS invocation log matrix ${\cal{Q}}$.

{\textbf{\em{(ii)}}} Once the user-service pairs are collected, the next task (refers to Block$_{32}$ of Fig. \ref{fig:architecture2}) is to identify the four QoS invocation log matrices corresponding to each user-service pair picked-up by the previous step.

{\textbf{\em{(iii)}}} The final task of the offline-learning module is to train NRegS$_2$ by the user-service pairs collected in the first sub-module of the offline-learning module, as discussed below:

\begin{itemize}
 \item For each user-service pair $(u_i, s_j)$, all four matrices containing $(u_i, s_j)$ are identified first.
 \item These four matrices are then used to train the four neural-networks of NRegS$_1$ to predict the QoS value of $s_j$ invoked by $u_i$ (refers to Block$_{41}$ of Fig. \ref{fig:architecture2}).
 \item Finally, the outputs obtained from NRegS$_1$ are served as a feature vector to train NRegS$_2$ targeting to generate ${\cal{Q}}[u_i, s_j]$ (refers to Block$_{42}$ of Fig. \ref{fig:architecture2}).
\end{itemize}

\noindent
In the next subsection, we discuss the online-learning module. 





\subsection{Online-learning Module}
\noindent
The online-learning module is executed once a query ${\cal{R}} = (u_{t_1}, s_{t_2})$ is raised. 
Fig. \ref{fig:architecture2} includes the pictorial overview of the online-learning module. 
Given a target user $u_{t_1}$ and a target service $s_{t_2}$, the first task of this module is to identify the four QoS invocation log matrices corresponding to $u_{t_1}, s_{t_2}$ (refers to Block$_{5}$ of Fig. \ref{fig:architecture2}). 
Once we have the four QoS invocation log matrices, we train the four neural-networks of NRegS$_1$ to predict the QoS value of $s_{t_2}$ to be invoked by $u_{t_1}$ (refers to Block$_{6}$ of Fig. \ref{fig:architecture2}). 
Finally, the predicted values obtained from NRegS$_1$ are fed to the pre-trained NRegS$_2$ to produce the final output (refers to Block$_{7}$ of Fig. \ref{fig:architecture2}).

In the next section, we discuss the performance of FES experimentally.

\section{Experimental Results}\label{sec:result}
\noindent
All experiments were performed on the MATLAB Online server version R2020b ({\url{https://matlab.mathworks.com}}) having the following configurations: Intel(R) Xeon(R) Platinum 8259CL CPU \@ 2.50GHz, x86\_64 architecture, 8 Cores(s), with cache configurations: L1d: 32K, L1i: 32K, L2: 1024K, L3: 36608K.}

\subsection{Datasets and Comparative Methods}
\noindent
We used 4 public datasets WS-DREAM-1 \cite{zheng2014investigating} and WS-DREAM-2 \cite{DBLP:conf/issre/ZhangZL11} to validate the performance of FES. The brief descriptions of the datasets are shown in Table \ref{tab:dataset}.

\begin{table}[!h]\makegapedcells
 \scriptsize
 \caption{Dataset description}
 \centering
 \begin{tabular}{c|c|c|c}
    \hlineB{2.5}
    \multirow{2}{*}{Dataset} & \multirow{2}{*}{QoS} & Dimension of          & 
Location\\
                             &                      & invocation log matrix & 
information\\
    \hlineB{2.5}
    \multirow{2}{*}{WS-DREAM-1} & RT & $339 \times 5825$ & Yes\\
    \cline{2-4}
    & TP & $339 \times 5825$ & Yes\\
    \hlineB{2.5}
    \multirow{2}{*}{WS-DREAM-2} & RT & $142 \times 4500 \times 64$ & No\\
    \cline{2-4}
    & TP & $142 \times 4500 \times 64$ & No\\
    \hlineB{2.5}
    \multicolumn{4}{c}{Dimension of invocation log matrix: \# users $\times$ \# services $\times$ \# timestamps}\\
    \multicolumn{4}{c}{RT: Response time; TP: Throughput}
 \end{tabular}\label{tab:dataset}
\end{table}

\noindent
It may be noted that WS-DREAM-2 dataset contains the user-service invocation log matrix for 64 timestamps. We performed our experiment on 64 individual matrices and recorded the average accuracy for each QoS parameter. Also, to show the performance of FES on the WS-DREAM-2 dataset, we skipped the context-aware clustering part (refers to FESWoCC), since WS-DREAM-2 dataset does not contain the contextual information of users or services.

\noindent
{\textbf{Training-testing Setup}}: In our experimental setup, we divided each dataset into three sets: training, validation, and testing. We performed our experiment for different training data sizes. If the size of the training data is considered to be $x\%$ of the total size of an invocation log matrix (${\cal{Q}}$), the remaining $(100 - x)\%$ of ${\cal{Q}}$ is divided into validation and testing set into a $1:2$ ratio. Each experiment was performed 5 times and the average results were recorded for analysis.

\noindent
{\textbf{Comparison Metric}}: To analyze the performance of FES, we used mean absolute error (MAE), which is calculated as the average of the absolute difference between the predicted and actual QoS values over the testing dataset.

\noindent
{\textbf{Comparative Methods}}: We compare FES with a set of state-of-the-art methods as presented in Table \ref{tab:soa}.

\begin{table}[!h]\makegapedcells
 \scriptsize
 \caption{Comparative Methods}
 \centering
 \begin{tabular}{l|l}
    \hlineB{2.5}    
    Type & Methods\\
    \hlineB{2.5}
    Collaborative filtering & UPCC \cite{breese1998empirical}, IPCC \cite{sarwar2001item}, WSRec \cite{zheng2011qos},  \\
    (Similarity-based) & NRCF \cite{sun2013personalized}, RACF \cite{wu2017collaborative}, RECF \cite{DBLP:conf/icsoc/ZouJNWPG18}\\\hline    
    Time series analysis & TA \cite{DBLP:conf/icws/AminCG12}\\\hline
    \multirow{2}{*}{Matrix factorization} & MF \cite{lo2012extended}, HDOP \cite{DBLP:journals/tsc/WangMCYC19}, NMF \cite{lee1999learning}, PMF \cite{mnih2008probabilistic},\\
    & NIMF \cite{DBLP:journals/tsc/ZhengMLK13}, WRAMF \cite{DBLP:journals/fgcs/ChenSYLS20}\\\hline
    Factorization Machine & EFM \cite{DBLP:conf/icsoc/WuXCCZ17}, AFM \cite{access/LiLZ20b}\\\hline
    Deep Architecture & DNM \cite{TSC8419326}, LDCF \cite{TSMC_8805172}, CADFM \cite{9189762}\\\hline
    Hybrid & CNR \cite{DBLP:conf/icsoc/ChattopadhyayB19}, CAHPHF \cite{chowdhury2020cahphf}, CAHPHFWoCF \cite{chowdhury2020cahphf}\\
    \hlineB{2.5}
 \end{tabular}\label{tab:soa}
\end{table}

\subsection{Configuration of FES}
\label{subsec:config}
\noindent
In this subsection, we discuss the configuration of hyper-parameters used for FES. We used two different configurations of FES in our experimental analysis.

\noindent
{\textbf{First Configuration of FES (FES-T)}}: In the first setup (say, FES-T), we transfer the configurations of the hyper-parameters from the implementation of CAHPHF. 

\noindent
{\textbf{Second Configuration of FES}}: In the second setup, we choose the configurations of the hyper-parameters of FES from the validation dataset. We now explain the details of the second-configuration.

\noindent
{\bf{Configuration of Multi-level Clustering}}: 
The multi-level user/service clustering combines 3 different clustering strategies. Altogether 12 different threshold parameters are required to be tuned in the clustering phase.
We introduce a data-driven parameter (say, $\uptau \in [0, 1)$) to determine the values of these thresholds. 

For \emph{context-aware clustering}, the threshold values were chosen as follows:
\begin{equation}\label{eq:tr_ca_u}\small
 T^u_c = \lfloor\uptau \times |{\cal{L}}^u(HD)|\rfloor^{th} \text{ element of }{\cal{L}}^u(HD)
\end{equation}
\begin{equation}\label{eq:tr_ca_s}\small
 T^s_c = \lfloor\uptau \times |{\cal{L}}^s(HD)|\rfloor^{th} \text{ element of }{\cal{L}}^s(HD)
\end{equation}
where ${\cal{L}}^u(HD)$ and ${\cal{L}}^s(HD)$ represents the sorted list (in ascending order) of Haversine distances between all pair of users and services, respectively.

For \emph{similarity-based clustering}, the threshold parameters were chosen as follows:
\begin{equation}\label{eq:tr_s_u}\small
 T^u_s = \lfloor\uptau \times |{\cal{L}}^u(CSM)|\rfloor^{th} \text{ element of }{\cal{L}}^u(CSM)
\end{equation}
\begin{equation}\label{eq:tr_s_s}\small
 T^s_s = \lfloor\uptau \times |{\cal{L}}^s(CSM)|\rfloor^{th} \text{ element of }{\cal{L}}^s(CSM)
\end{equation}
\noindent
where ${\cal{L}}^u(CSM)$ and ${\cal{L}}^s(CSM)$ represent the sorted list (in descending order) of all-pair cosine similarity measures across users and services, respectively.

For \emph{context-sensitivity-based clustering}, the values of the thresholds are determined as follows:
\begin{equation}\label{eq:tr_cs_u}\small
 T^u_{cs} = max(N_u, \uptau \times |One{\cal{UCL}}_c|)
\end{equation}
\begin{equation}\label{eq:tr_cs_s}\small
 T^s_{cs} = max(N_s, \uptau \times |One{\cal{SCL}}_c|)
\end{equation}

\noindent
Empirically, $\uptau$ was set to 0.5 (referred to Fig. \ref{fig:coldStart}:(b)), whereas both $N_u$ and $N_s$ were chosen as 100 for WS-DREAM-1, and 65 for WS-DREAM-2.

\noindent
{\bf{Configuration of NRegS$_1$}}:
NRegS$_1$ consists of 4 neural-networks. Since NRegS$_1$ is trained in online-mode, we aim to spent as less-time as possible to train the neural networks. Each neural network comprised 2 hidden layers containing 32 and 16 neurons, respectively. 
Here, we used the sigmoid activation function for the hidden layers and the linear activation function for the output layer \cite{haykin2009neural}.  
For optimizing the learning parameters, we used stochastic gradient descent with momentum \cite{haykin2009neural}. The learning rate of each network was set to 0.01 with a momentum of 0.9. The mean squared error was used as the cost function \cite{haykin2009neural}. The termination criteria of each network were set to either up to a minimum gradient of $10^{-5}$ or up to 50 epochs.

\noindent
{\bf{Configuration of NRegS$_2$}}: 
NRegS$_2$ comprises a single neural network consisting of only 2 hidden layers with 8 neurons each. 
Similar to NRegS$_1$, here also, the sigmoid activation function was used for the hidden layers and the linear activation function was used for the output layer \cite{haykin2009neural}. 
To optimize the learning parameters, we used mini-batch gradient descent with momentum \cite{haykin2009neural} with a batch size of 100. The learning rate of the network was fixed to 0.01 with a momentum of 0.9. Here also, the mean squared error was used as the cost function \cite{haykin2009neural}. The training was conducted up to a minimum gradient of $10^{-7}$ or up to 5000 epochs. The size of the training dataset (say, $|TrD_{S2}|$) to train NRegS$_2$ varied between 1800 to 2200.

In Section \ref{subsec:impact}, we show the impact of the hyper-parameters on the performance of FES.

\subsection{Experimental Analysis}
\noindent
In this section, we present the experimental analysis of FES to establish the three major goals of this work. 

\subsubsection{Accuracy of FES}
\noindent
Tables \ref{tab:compareDS1} and \ref{tab:compareDS2} present the comparison among FES (with the first and second configurations) and the major SoA methods in terms of prediction accuracy (measured by MAE) on WS-DREAM-1 and WS-DREAM-2 datasets, respectively. 

\begin{table}[!h]\makegapedcells
\scriptsize
\caption{Performance (MAE) of FES and comparison with various SoA methods over WS-DREAM-1 \cite{zheng2014investigating}}
\centering
\begin{minipage}{.5\linewidth}
 \begin{tabular}{c|c|c} 
 \hline
 \multicolumn{3}{c}{QoS: RT}\\
 \hline
 \multirow{2}{*}{Method} &  \multicolumn{2}{c}{$|TrD|$} \\
 \cline{2-3}
 & 10\% & 20\% \\
 \hline\hline
 UPCC & 0.6063 & 0.5379\\
 \hline
 IPCC & 0.7000 & 0.5351\\
 \hline
 WSRec & 0.6394 & 0.5024\\
 \hline
 MF & 0.5103 & 0.4981 \\
 \hline
 NRCF & 0.5312 & 0.4607\\
 \hline
 RACF & 0.4937 & 0. 4208\\
 \hline
 NIMF & 0.4854 & 0.4357\\
 \hline
 AFM & 0.4849 & 0.3847\\
 \hline
 WRAMF & 0.4657 & - \\
 \hline
 EFM & 0.4446 & - \\
 \hline
 RECF & 0.4332 & 0.3946\\
 \hline
 CADFM & 0.3896 & 0.3600\\
 \hline
 LDCF & 0.3670 & 0.3310\\
 \hline
 DNM & 0.3628 & -\\
 \hline
 CNR & 0.2597 & 0.1711\\
 \hline
 CAHPHF & 0.0590 & 0.0419\\
 \hline
 {\textbf{FES-T}} & {\textbf{0.0706}} & {\textbf{0.0691}}\\
 \hline
 {\textbf{FES}} & {\textbf{0.0349}} & {\textbf{0.0247}}\\
 
 \hline
 \end{tabular}
 \end{minipage}%
 \begin{minipage}{.5\linewidth}
   \begin{tabular}{c|c|c} 
 \hline
 \multicolumn{3}{c}{QoS: TP}\\
 \hline
 \multirow{2}{*}{Method} &  \multicolumn{2}{c}{$|TrD|$} \\
 \cline{2-3}
 & 10\% & 20\% \\
 \hline\hline
 UPCC & 21.2695 & 17.5546\\
 \hline
 IPCC & 27.3993 & 25.0273\\
 \hline
 WSRec & 19.9754 & 16.0762\\
 \hline
 NMF & 17.8411 & 15.2516\\
 \hline
 PMF & 16.1755 & 14.6694\\
 \hline
 NIMF & 16.0542 & 13.7099\\
 \hline
 EFM & 15.0268 & -\\
 \hline
 DNM & 12.9200 & -\\
 \hline
 LDCF & 12.3820 & 10.8410\\
 \hline
 CAHPHF & 5.9800 & 4.1890\\
 \hline
 {\textbf{FES-T}} & {\textbf{11.0100}} & {\textbf{10.7000}}\\
 \hline
 {\textbf{FES}} & {\textbf{4.5490}} & {\textbf{4.0090}}\\
 \hline
 \multicolumn{3}{r}{$|TrD|$: Size of Training Data} \\
 \end{tabular}
 \end{minipage}

 \label{tab:compareDS1}
\end{table}

\begin{table}[!h]\makegapedcells
\scriptsize
\caption{Comparison (MAE) of FES with various SoA methods over WS-DREAM-2 \cite{DBLP:conf/issre/ZhangZL11}}
\centering
 \begin{tabular}{c|c|c|c|c} 
 \hline
 \multirow{3}{*}{Method} & \multicolumn{2}{c|}{QoS: RT} & \multicolumn{2}{c}{QoS: TP}\\
 \cline{2-5}
 &  \multicolumn{2}{c|}{$|TrD|$} &  \multicolumn{2}{c}{$|TrD|$}\\
 \cline{2-5}
 & 10\% & 20\% & 10\% & 20\% \\
 \hline\hline
 MF & 0.4987 & 0.4495 & 16.3214 & 14.1478 \\
 \hline
 TA & 0.6013 & 0.5994 & 17.2365 & 15.0994 \\
 \hline
 HDOP & 0.3076 & 0.2276 & 13.2578 & 10.1276 \\
 \hline
 CAHPHFWoCF & 0.1187 & 0.0758 & 4.8970 & 4.1010 \\
 \hline
 {\textbf{FESWoCC-T}} & {\textbf{0.1221}} & {\textbf{0.0940}} & {\textbf{5.0700}} & {\textbf{4.7060}} \\
 \hline
 {\textbf{FESWoCC}} & {\textbf{0.0988}} & {\textbf{0.0704}} & {\textbf{4.1660}} & {\textbf{3.7760}} \\ 
 \hline
 \multicolumn{5}{r}{$|TrD|$: Size of Training Data}
\end{tabular}\label{tab:compareDS2}
\end{table}

\noindent
From Tables \ref{tab:compareDS1} and \ref{tab:compareDS2}, we have the following observations: 

\textbullet~ {\emph{Performance of FES:}} As evident from Tables \ref{tab:compareDS1} and \ref{tab:compareDS2}, FES performed the best as compared to the contemporary methods for both the QoS parameters on both the datasets. On average, FES achieved a 19.6\% improvement as compared to the second-best method (i.e., CAHPHF/CAHPHFWoCF) enlisted in Tables \ref{tab:compareDS1} and \ref{tab:compareDS2}.
 
\textbullet~ {\emph{Impact of the cardinality of the training dataset:}} As observed from the experimental study on the four different datasets shown in Tables \ref{tab:compareDS1} and \ref{tab:compareDS2}, the MAE value reduced with the increase in the cardinality of the training datasets.
 
\textbullet~ {\emph{Comparing FES-T with CAHPHF:}} As we discussed earlier, the first configuration of FES (i.e., FES-T) was adopted from the implementation of CAHPHF. 
It can be noted from Tables \ref{tab:compareDS1} and \ref{tab:compareDS2}, CAHPHF performed better than FES-T in terms of prediction accuracy. One possible reason is that the training of second-level neural-network in CAHPHF focuses only on one set of users and services containing the target user and service, whereas, in FES-T, the training of NRegS$_2$ is performed across all the clusters. Therefore, NRegS$_2$ needs to be trained properly to avoid underfitting \cite{haykin2009neural}. However, FES is better than CAHPHF in terms of prediction accuracy. 
%

\begin{table}[!h]\makegapedcells
\scriptsize
\caption{Comparison between CAHPHF with FES in terms of single prediction time (in sec)}
\centering
 \begin{tabular}{c|c|c|c|c}
 \hlineB{2.5}
 Dataset & QoS & $|TrD|$ & CAHPHF & FES \\
 \hlineB{2.5}
 \multirow{4}{*}{WS-DREAM-1} & \multirow{2}{*}{RT} & 10\% & 312.6 & 0.017 \\
 \cline{3-5}
 & & 20\% & 319.8 & 0.019 \\
 \cline{2-5}
 & \multirow{2}{*}{TP} & 10\% & 357.8 & 0.019\\
 \cline{3-5}
 & & 20\% & 374.6 & 0.022\\
 \hlineB{2.5}
  &  &  & CAHPHFWoCF & FESWoCC \\
 \hlineB{2.5}
 \multirow{4}{*}{WS-DREAM-2} & \multirow{2}{*}{RT} & 10\% & 413.4 & 0.044 \\
 \cline{3-5}
 & & 20\% & 438.9 & 0.052 \\
 \cline{2-5}
 & \multirow{2}{*}{TP} & 10\% & 467.3 & 0.049\\
 \cline{3-5}
 & & 20\% & 488.2 & 0.057\\
 \hlineB{2.5}
\end{tabular}\label{tab:compareCT}
\end{table}

\subsubsection{Responsiveness of FES}
\noindent
Table \ref{tab:compareCT} presents the comparison between FES and CAHPHF in terms of the prediction time (also referred to as the responsiveness) on both datasets.
Fig. \ref{fig:ct}:(a) shows the performance of FES in terms of the prediction time on the WS-DREAM-1:RT dataset. From now onwards, we present our experimental study on the WS-DREAM-1:RT dataset since similar trends are observed on the other datasets. 
From Table \ref{tab:compareCT} and Fig. \ref{fig:ct}:(a), we have the following observations:

\textbullet~ {\emph{Comparison between FES and CAHPHF}}: As evident from Table \ref{tab:compareCT}, FES is significantly faster than CAHPHF, due to its preprocessing module and semi-offline training strategy.

\textbullet~ {\emph{Performance of FES}}: As observed from Fig. \ref{fig:ct}:(a), FES is faster than the contemporary methods in terms of prediction time. Here, we consider the SoA methods which are known for their faster responsiveness. On average FES achieved a 6.9x speed-up to predict a single value as compared to the second-best method reported in Fig. \ref{fig:ct}:(a).

\begin{figure}[!h]
\centering
(a) \includegraphics[width=0.4\linewidth]{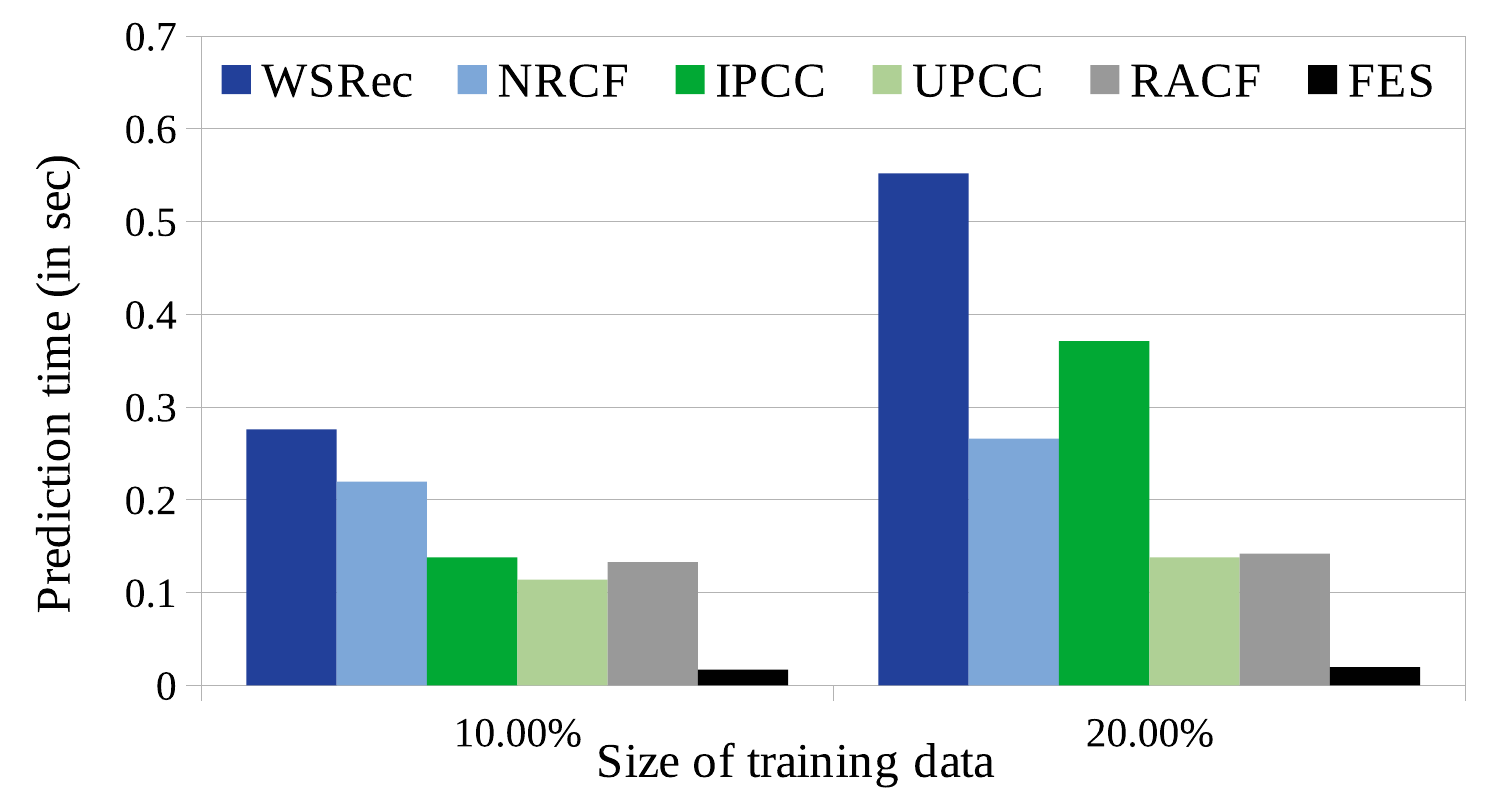}
(b) \includegraphics[width=0.4\linewidth]{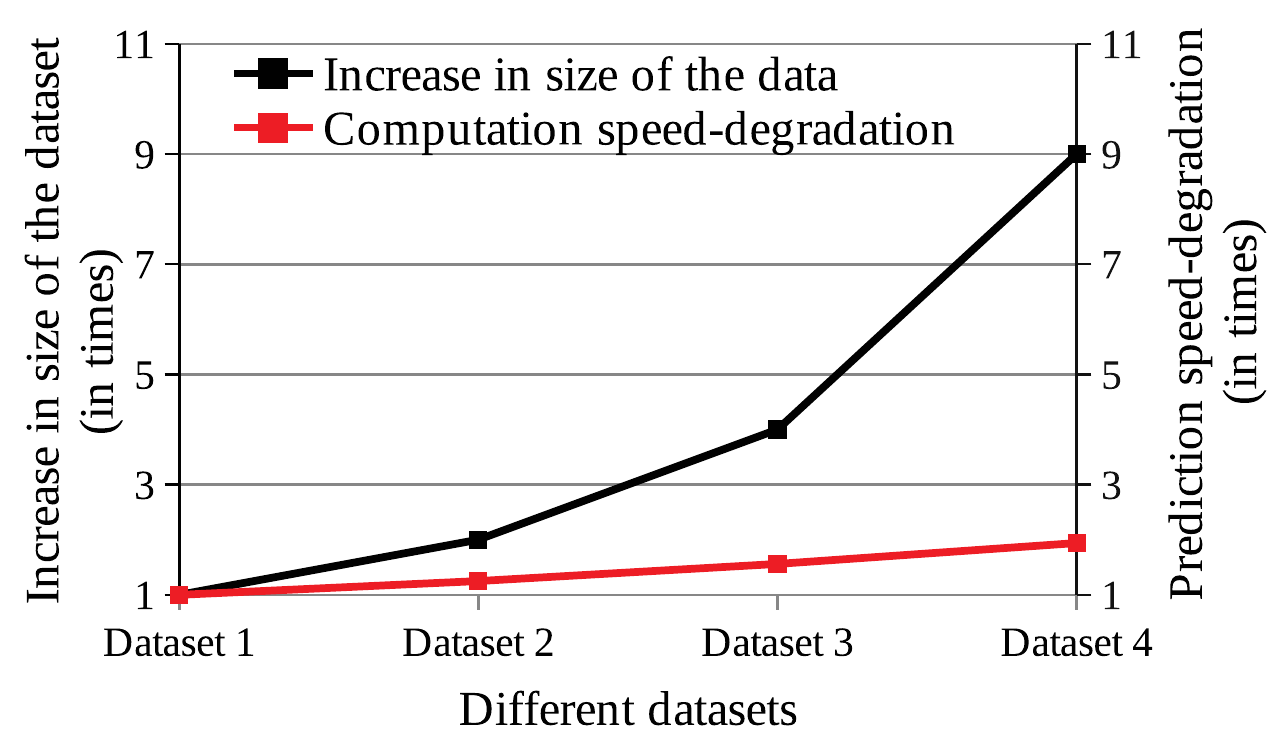}
\caption{(a) Responsiveness analysis of FES; (b) Scalability analysis of FES}
\label{fig:ct}
\end{figure}

%

\subsubsection{Scalability of FES}
\noindent
To show the scalability of FES, we created four augmented datasets using WS-DREAM-2:RT dataset, as shown in Table \ref{tab:augData}.
The detailed description can be found in \cite{chowdhury2020cahphf}.


Fig. \ref{fig:ct}:(b) shows the relative performance of FES in terms of scalability on the augmented datasets  
with respect to the same on the first matrix of the WS-DREAM-2:RT dataset (say, Dataset 1).
The primary $y$-axis of the figure shows the increase in the size of the dataset, which is measured as the ratio between the size of the augmented dataset and the size of Dataset 1. The secondary $y$-axis, on the other hand, shows the prediction time degradation by FES, which is measured as the ratio between the prediction time of FES on Dataset 1 and the prediction time of FES on Dataset $i$, for $i = \{2, 3, 4\}$.
It may be noted that Dataset~2 contains two datasets with different cardinalities, for which we recorded the average prediction time degradation.
As observed from Fig. \ref{fig:ct}:(b), the prediction time increases at a significantly slower rate as compared to the growth of the dataset-cardinality, which implies that FES is highly scalable.

\begin{table}[!t]\makegapedcells
\scriptsize
\caption{Brief details of augmented datasets}
\centering
 \begin{tabular}{c|c|c} 
\hline
Dataset & Size (\# users $\times$ \# services) & Increment (times) \\ \hline \hline
Dataset~1  & \multirow{2}{*}{$142 \times 4500$} & \multirow{2}{*}{1x} \\ 
(WS-DREAM-2:RT) &  &  \\ \hline
\multirow{2}{*}{Dataset~2} & $284 \times 4500$ & \multirow{2}{*}{2x} \\ \cline{2-2} 
& $142 \times 9000$ &  \\ \hline
Dataset~3 & $284 \times 9000$ & 4x \\ \hline
Dataset~4 & $426 \times 13500$ & 9x \\ \hline
\end{tabular}\label{tab:augData}
\end{table}

\subsubsection{Tackling Cold-Start by FES}
\noindent
We now analyze how FES deals with the cold-start situation. The cold-start is a scenario when new users/services are added to the system. In general, for cold-start, the prediction accuracy drops considerably since the newly added users/services do not have any record. However, FES deals with cold-start by performing contextual clustering on newly-added users/services.

To show the performance of FES in the cold-start situation, we created a dataset for our experiment. In the QoS log matrix, 25\% of users (services) data were replaced by 0, representing the newly added users (services).
Fig. \ref{fig:coldStart}:(a) shows the performance of FES in the cold-start scenario. In the figure, 4* represents, 25\% of users  (services) data were replaced by 0 in the QoS invocation log matrix, while the other users (services) have only 4\% data available.
As observed from Fig. \ref{fig:coldStart}:(a), in the cold-start situation, the prediction accuracy of FES is reasonably good.

\begin{figure}[!h]
\centering
(a) \includegraphics[width=0.4\linewidth]{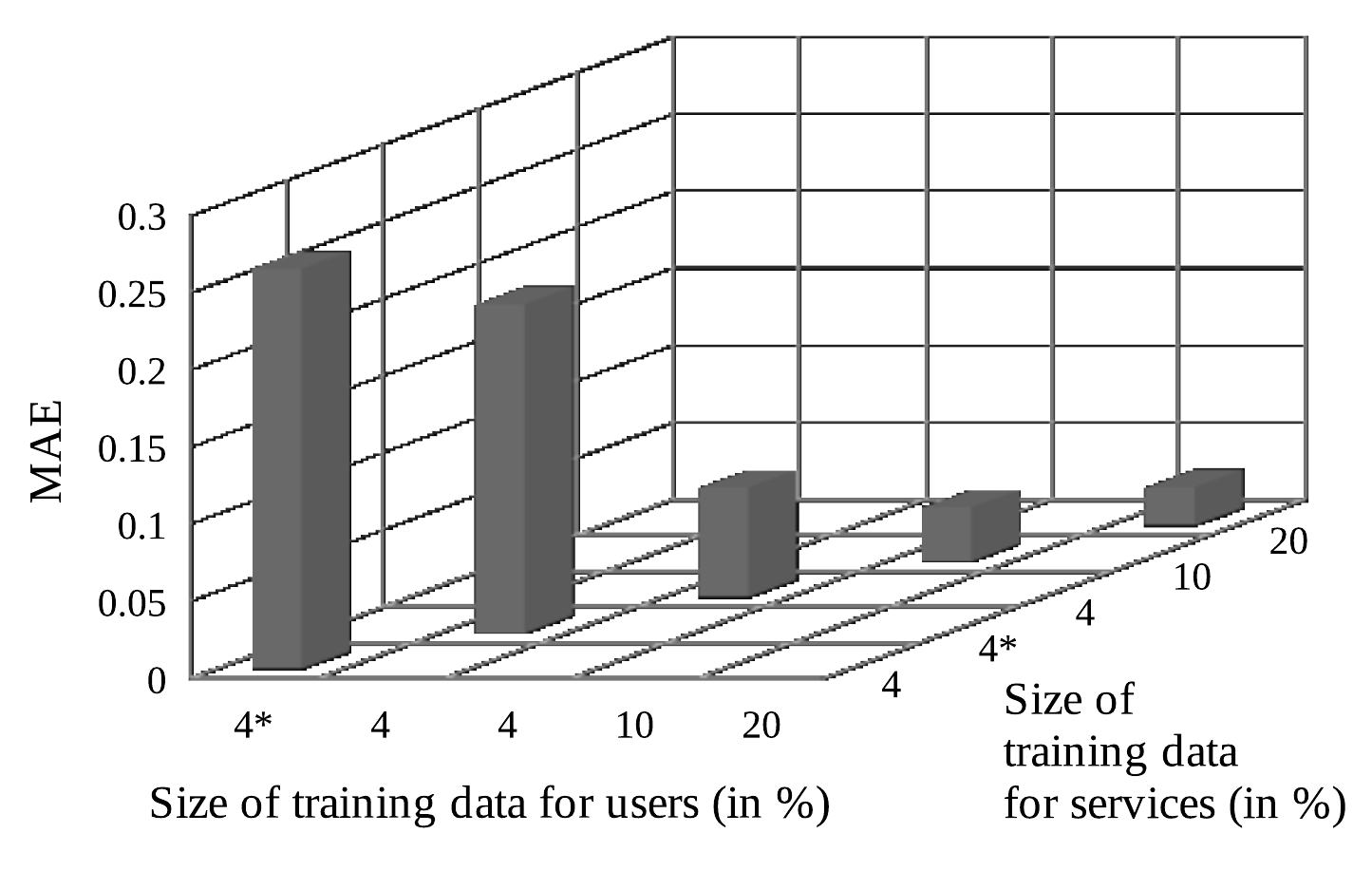}
(b) \includegraphics[width=0.4\linewidth]{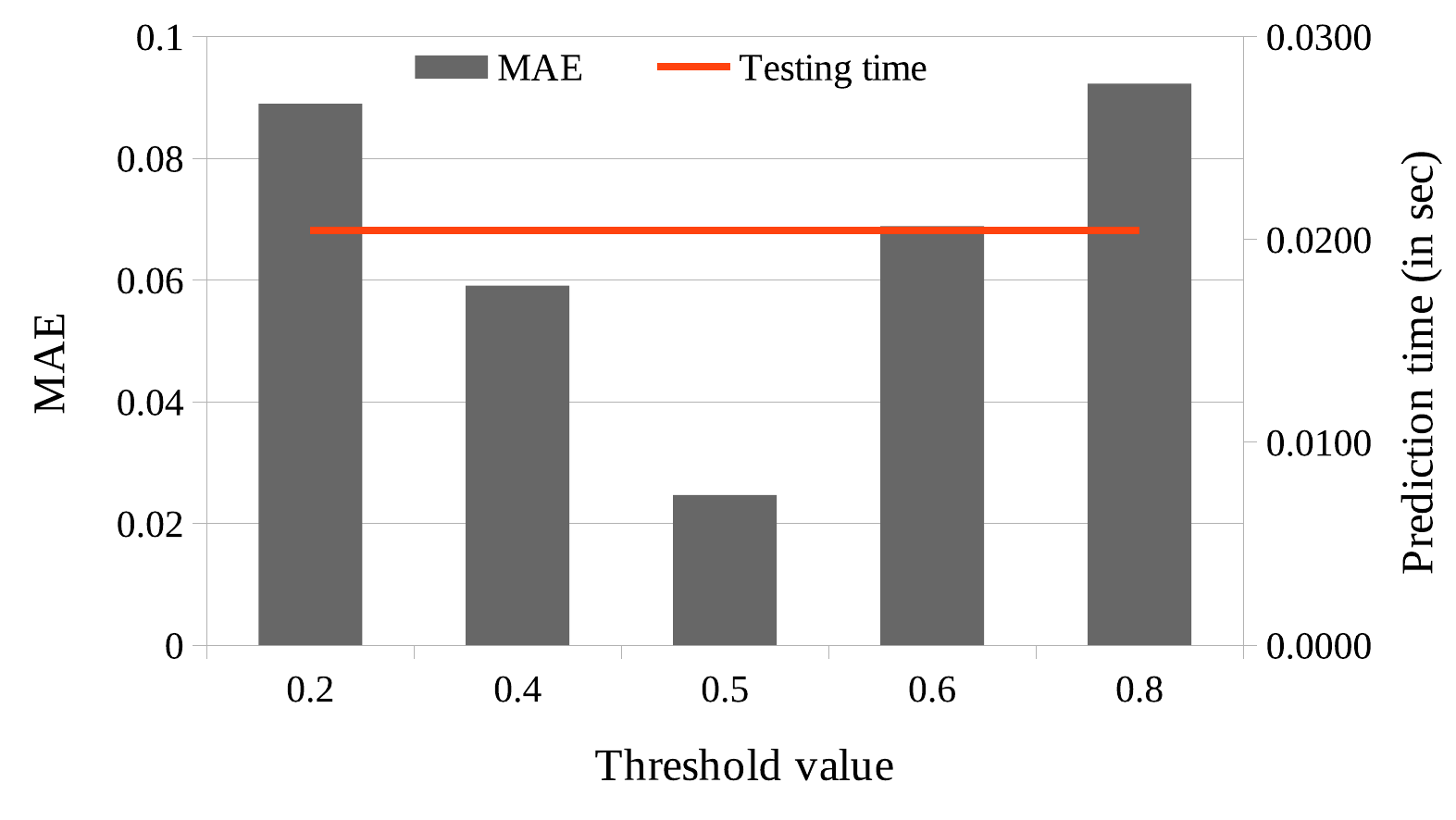}
\caption{(a) Analysis for cold-start; (b) Impact of the threshold used for clustering}
\label{fig:coldStart}
\end{figure}

\subsubsection{Impact of Hyperparameters}\label{subsec:impact}
\noindent
In this section, we discuss the impact of the hyperparameters on the 
performance of FES. 

\medskip
\noindent
a) {\emph{Impact of Threshold $\uptau$ in Multi-level Clustering}}: Fig. \ref{fig:coldStart}:(b) shows the impact of the threshold $\uptau$ introduced in Section \ref{subsec:config}. It may be noted that $\uptau$ controls the number of clusters and as well as, the number of users/services in a cluster. As evident from the figure, for $\uptau = 0.5$, FES performed the best. The prediction accuracy deteriorated with the increase/decrease in the value of $\uptau$ beyond 0.5. 
When the value of $\uptau$ decreased beyond 0.5, the MAE increased (i.e., prediction accuracy decreased). This is because the number of users/services in a cluster decreased with the decrease in the value of $\uptau$. This, in turn, introduced the underfitting problem due to the insufficient training data for NRegS$_1$. 
On the other hand, when the value of $\uptau$ increased beyond 0.5, the MAE increased because of more number of users/services in a cluster, and thereby, less correlated data in the training dataset of NRegS$_1$.

%
\medskip
\noindent
b) {\emph{Impact of Training Data Size of NRegS$_2$}}: Fig. \ref{fig:trdNRegS2}:(a) shows the performance of FES with respect to the size of the training dataset of NRegS$_2$. As evident from Fig. \ref{fig:trdNRegS2}:(a), the value of MAE decreases with the increase in the size of the training dataset of NRegS$_2$. However, the size of the training dataset of NRegS$_2$ does not have any impact on the prediction time.

\begin{figure}[!h]
\centering
(a) \includegraphics[width=0.4\linewidth]{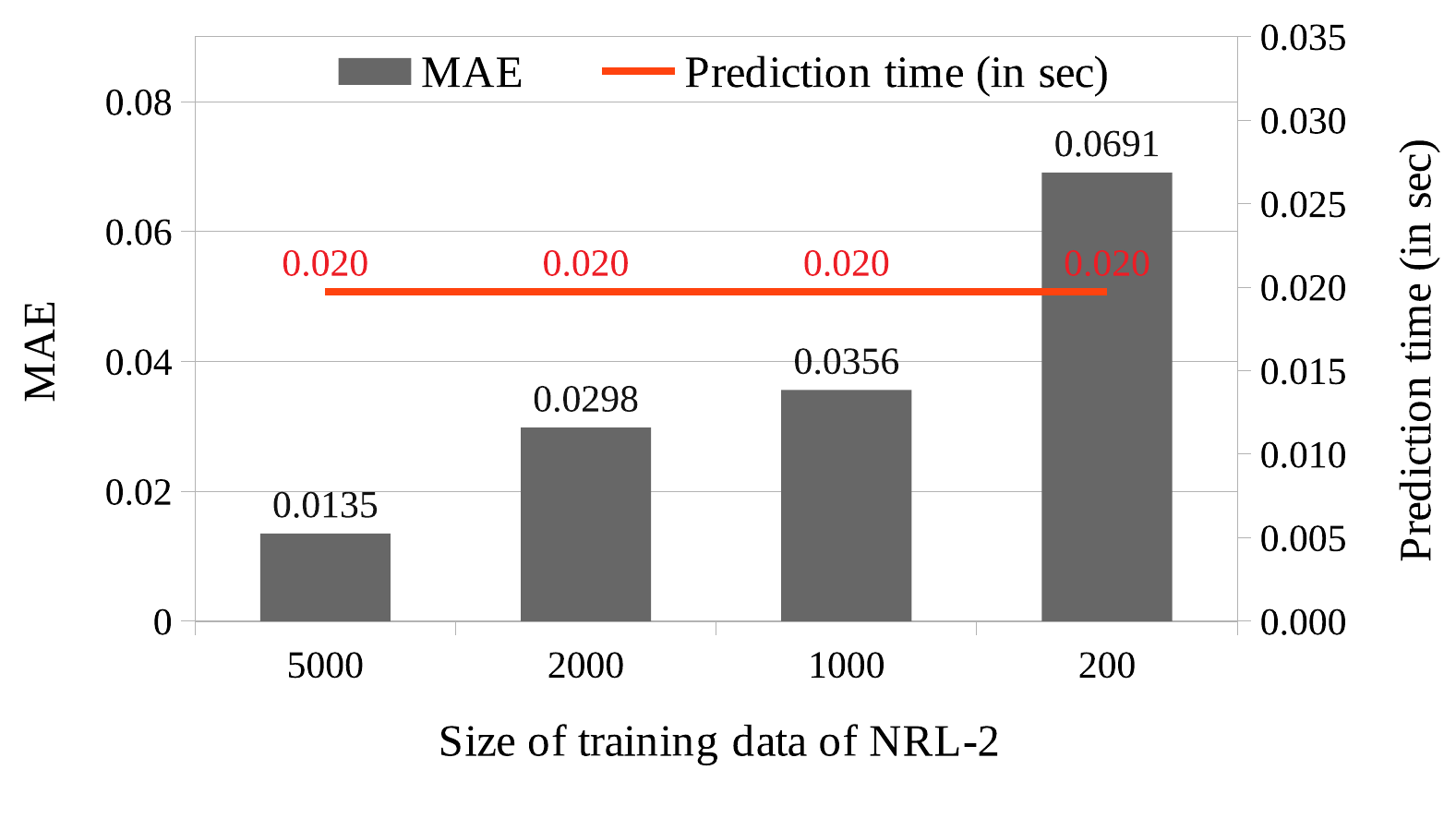}
(b) \includegraphics[width=0.4\linewidth]{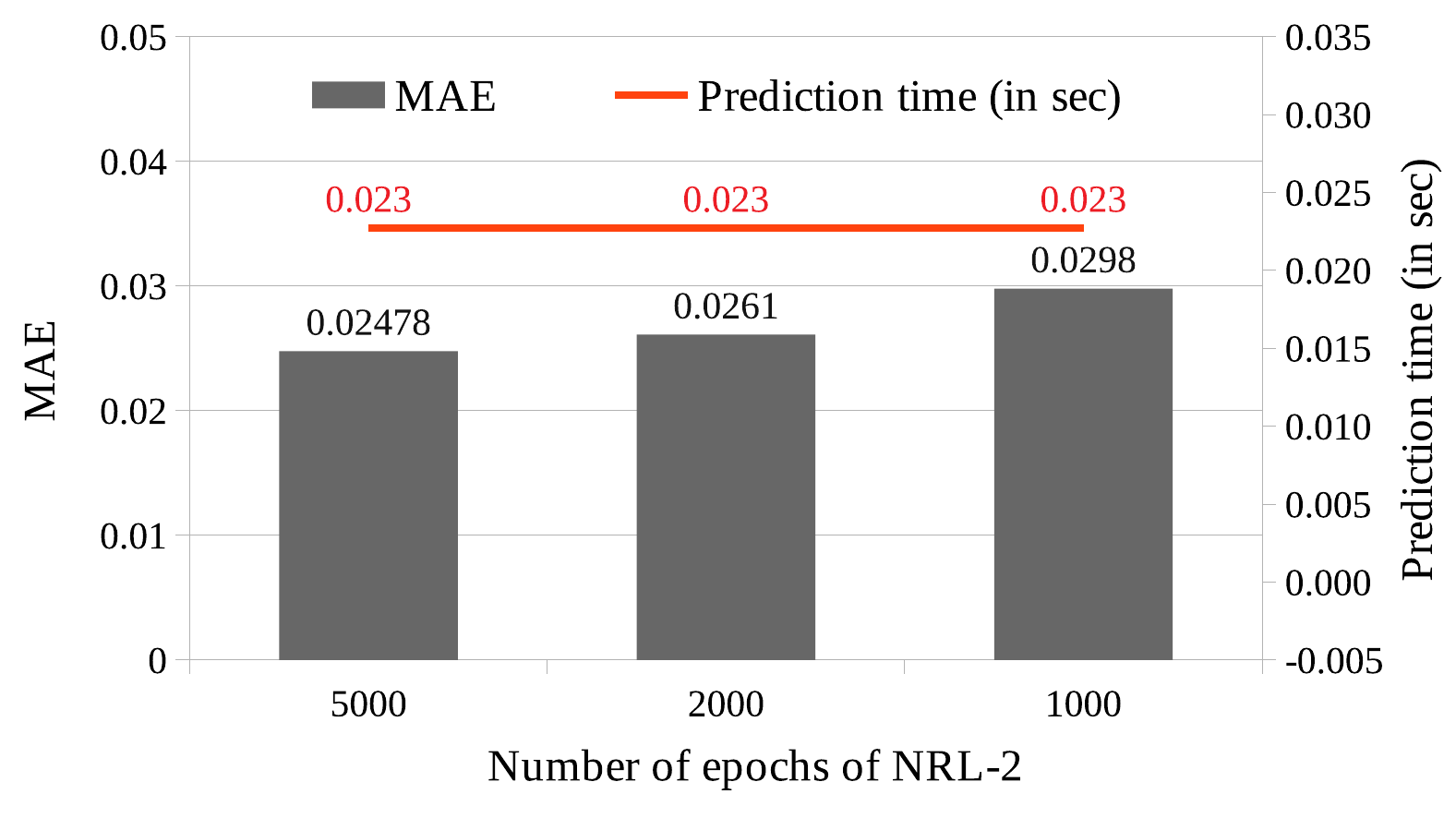}
\caption{Impact of (a) training data size of NRegS$_2$, (b) the number of epochs of NRegS$_2$}
\label{fig:trdNRegS2}
\end{figure} 

\noindent
c) {\emph{Impact of Number of Epochs of NRegS$_2$}}: Fig. \ref{fig:trdNRegS2}:(b) shows the performance of FES with respect to the 
number of epochs of NRegS$_2$. As observed from Fig. \ref{fig:trdNRegS2}:(b), the value of MAE decreases with the increase in the number of epochs of NRegS$_2$. However, similar to the previous hyperparameter, the number of epochs of NRegS$_2$ does not have any impact on the prediction time, as evident from this figure. 

%
\medskip
\noindent
d) {\emph{Impact of Number of Epochs of NRegS$_1$}}: Fig. \ref{fig:epochNRegS1}:(a) shows the performance of FES with respect to the 
number of epochs of NRegS$_1$. As observed from Fig. \ref{fig:epochNRegS1}:(a), the value of MAE decreases with the increase in the number of epochs of NRegS$_1$. However, unlike the previous hyperparameters, the number of epochs of NRegS$_1$ has an impact on the prediction time since the NRegS$_1$ is trained online, more specifically, at real-time prediction. Therefore, in the configuration of FES, we keep the number of epochs of NRegS$_1$ as less as possible. 

\begin{figure}[!h]
\centering
(a) \includegraphics[width=0.4\linewidth]{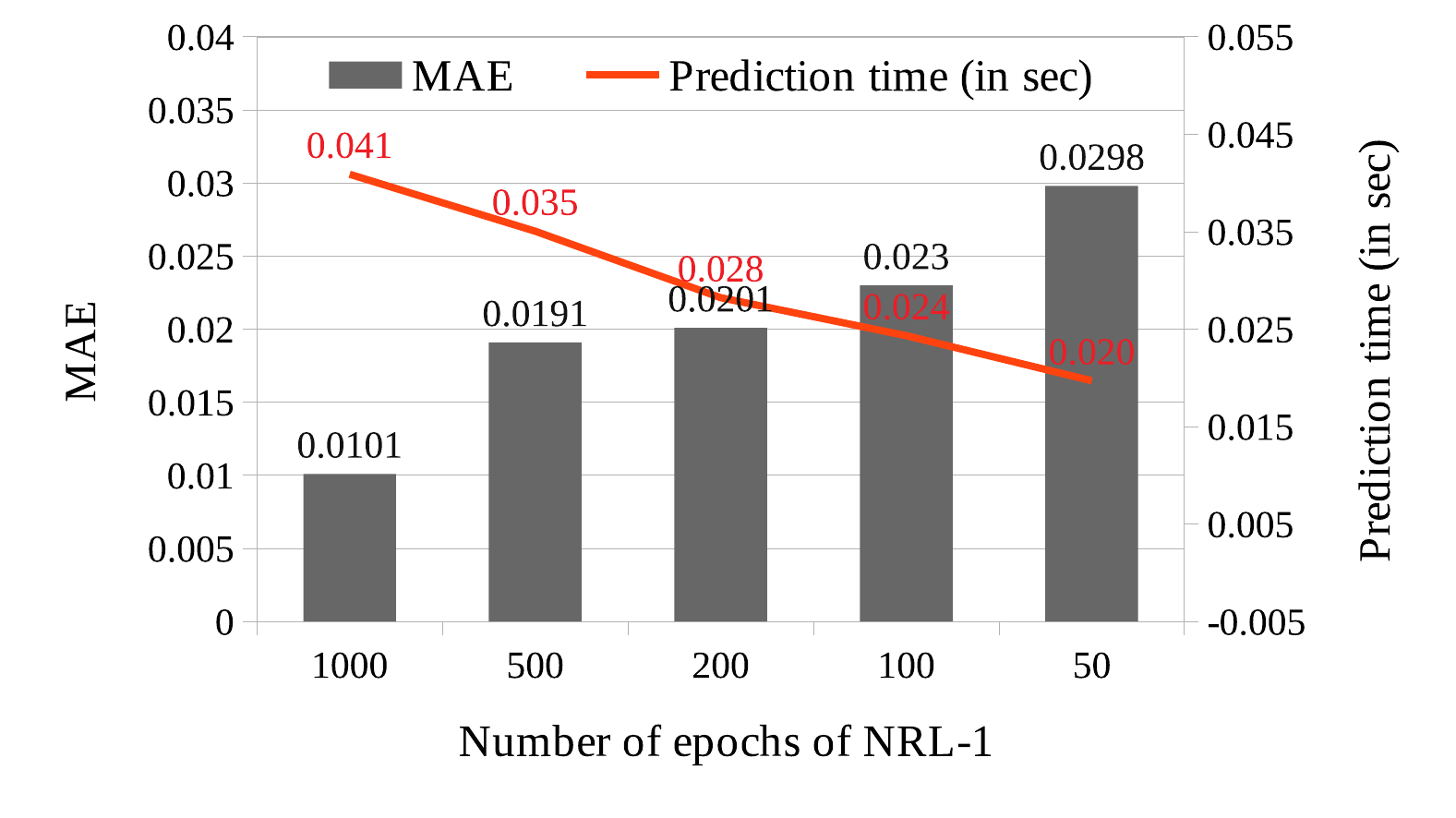}
(b) \includegraphics[width=0.4\linewidth]{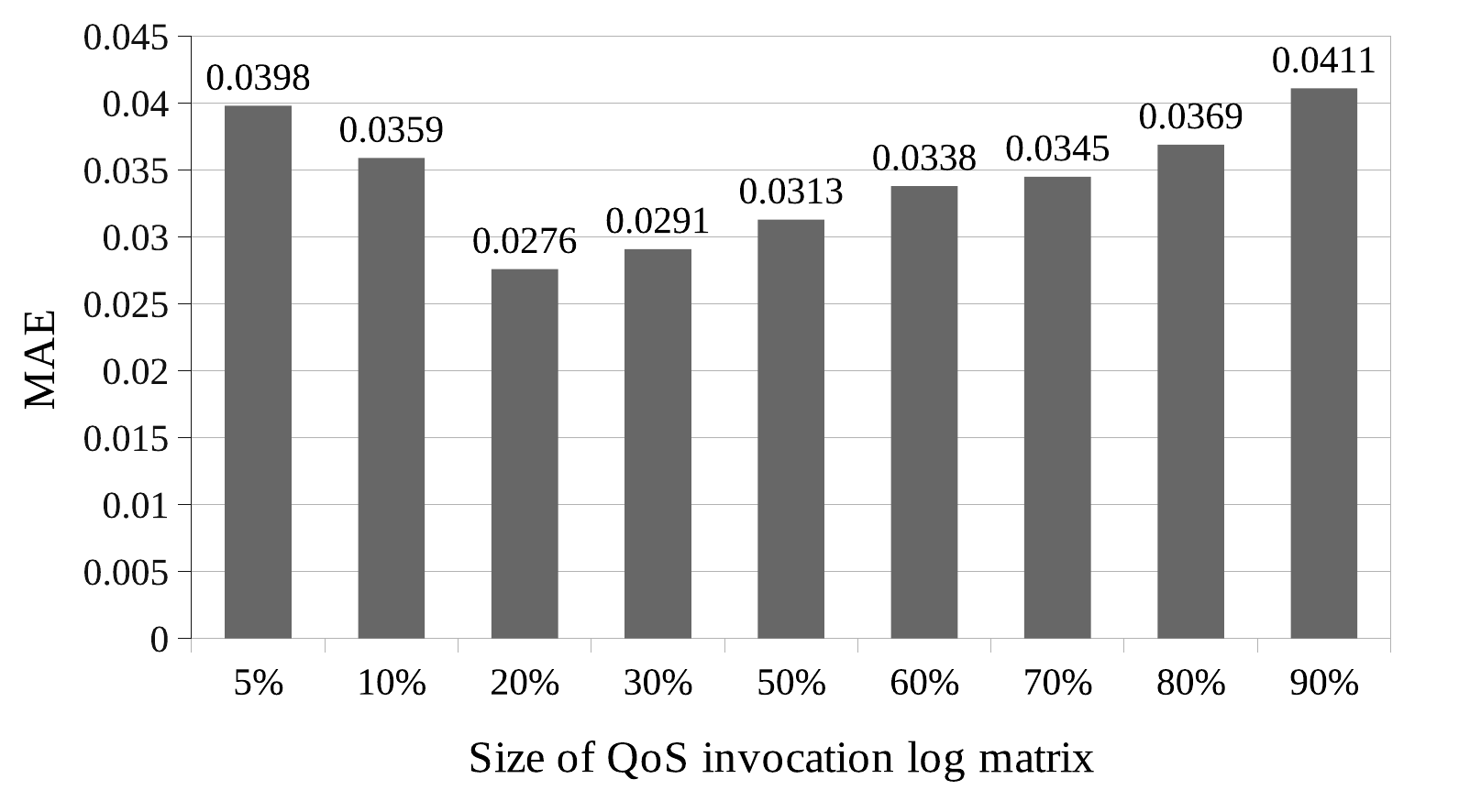}
\caption{Impact of (a) the number of epochs of NRegS$_1$, (b) semi-offline training}
\label{fig:epochNRegS1}
\end{figure} 

%

\subsubsection{Impact of Semi-offline Training}
\noindent
As discussed earlier, FES involves two modes of training:  
\begin{itemize}
 \item Offline training for NRegS$_2$: This requires training NRegS$_1$ multiple times (precisely, $|TrD_{S2}|$ times) to generate each training sample for NRegS$_2$.
 \item Online training for NRegS$_1$: NRegS$_1$ is required to be trained once again in the online mode to generate the test sample for NRegS$_2$ for the target QoS prediction.
\end{itemize}

\begin{figure}[!h]
\centering
\includegraphics[width=\linewidth]{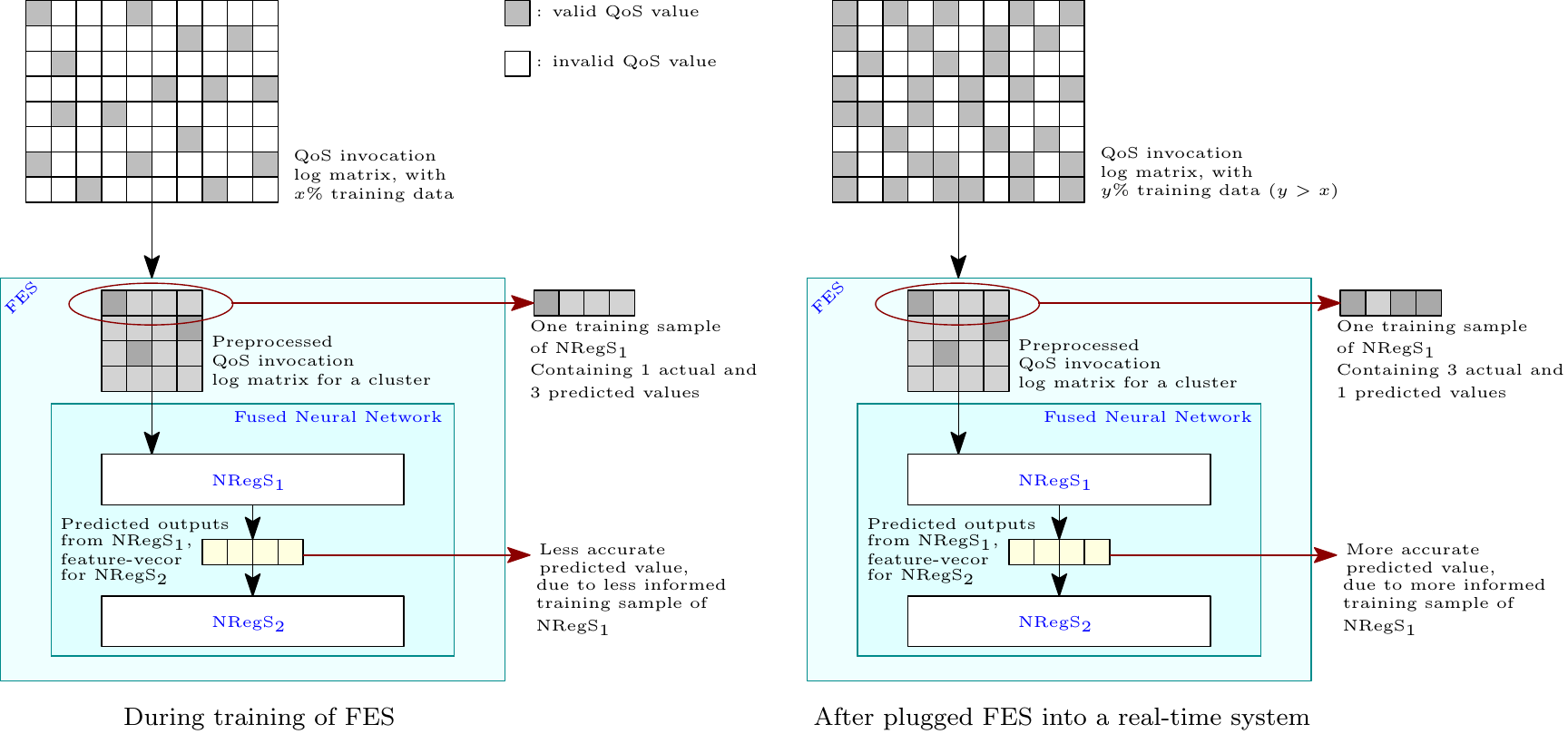}
\caption{Training snippet of FES and fused neural network}
\label{fig:trainingSnippet}
\end{figure} 

\noindent
Before discussing the impact of semi-offline training, we first discuss a few details of the training data samples for NRegS$_1$.
It may be noted that the number of valid entries of the QoS invocation log matrix ${\cal{Q}}$ starts increasing once FES is deployed as part of a real-time system due to the continuous addition of the actual QoS value of a new user-service pair in ${\cal{Q}}$.
Initially when the framework is just deployed, immediately after the training of FES, the model accuracy is reflected by the performance of FES in real-time. However, after some time, when ${\cal{Q}}$ is filled-up by more number of actual QoS values, the training sample of NRegS$_1$ becomes more informed, as depicted in Fig. \ref{fig:trainingSnippet}. As a result, the prediction accuracy of NRegS$_1$ is enhanced. However, due to the change in prediction accuracy of NRegS$_1$, the sample distribution of NRegS$_2$ is changed, and thereby, the prediction accuracy of NRegS$_2$ drops. Consequently, the prediction accuracy of FES drops eventually. This leads to  the necessity of retraining of FES.

Here, we conducted another experiment to show the impact of the semi-offline training of our framework on the long-term prediction accuracy of FES. Fig. \ref{fig:epochNRegS1}:(b) shows the impact on the prediction accuracy of FES due to the growth of ${\cal{Q}}$.
In this experiment, the FES was trained on 5\% data in ${\cal{Q}}$. The $x$-axis of Fig. \ref{fig:epochNRegS1}:(b) shows the increasing growth of the QoS invocation log matrix after deployment of our framework, whereas, the $y$-axis of the figure shows the MAE obtained by FES. From the figure, the followings have been observed:

\textbullet~{\emph{Impact of offline-online training}}: With the increase in the number of valid entries in ${\cal{Q}}$, the MAE of FES first decreased and then started increasing after a certain point. One possible reason for this trend is as follows. The MAE of FES at run-time depends on the accuracy of NRegS$_1$ and NRegS$_2$ simultaneously.
 With the increase in the number of valid entries in ${\cal{Q}}$, the accuracy of NRegS$_1$ increases, but the accuracy of NRegS$_2$ decreases. At the beginning of the experiment, the overall MAE of FES decreased possibly because the accuracy improvement of NRegS$_1$ dominated the accuracy degradation of NRegS$_2$. However, after a certain point, the accuracy degradation of NRegS$_2$ became the dominating factor. Hence, the overall MAE of FES started increasing.

\textbullet~{\emph{Requirement of retraining}}: It may be noted that in the case of (5\%, 5\%) data (where the first element of the tuple refers to the percentage of valid entries in ${\cal{Q}}$ during the training of FES, and the second element refers to the percentage of valid entries in ${\cal{Q}}$ after addition of more QoS records in run-time) FES achieved an MAE of 0.0398, which may be considered as an acceptable prediction error. 
 For (5\%, 10\%) and (5\%, 20\%) cases, although the MAE (0.359 and 0.0276, respectively) decreased with respect to (5\%, 5\%) case, they were still more than the MAE achieved by FES in (10\%, 10\%) and (20\%, 20\%) cases (0.0349 and 0.247, respectively).
 This implies, at any point in time, if we retrain FES, we can obtain better prediction accuracy. 
 However, since the prediction accuracy obtained by FES in (5\%, x\%) cases, where x $\in$ [5, 80], were better than the acceptable prediction accuracy, we may avoid further training of FES.
 In the case of (5\%, 90\%), FES attained an MAE of 0.0411, which was more than the acceptable prediction error. Therefore, at this point, retraining of FES is recommended. 

In summary, we can infer from our experimental results that FES is scalable, fast, and efficient (in terms of prediction accuracy) as compared to the SoA methods.

\section{Conclusion}
\label{sec:conclusion}
\noindent
In this paper, we propose a semi-offline QoS prediction framework by leveraging multi-level clustering and multi-phase prediction to ensure higher accuracy, faster responsiveness, and high scalability.
The experimental analysis on four large-scale benchmark WS-DREAM datasets shows that our framework outperformed the state-of-the-art methods in terms of prediction accuracy, prediction time, and scalability. 
Although in this paper, we exploit location information of users/services to enhance the prediction accuracy, the QoS variation overtime is out of the scope of our current research.
Furthermore, the other contextual information, such as network parameters, the variation of the QoS parameters based on different execution platforms of the services, are also not considered in this framework.
Our future endeavor involves working on time-aware QoS prediction along with exploring other contextual information.

\balance

\bibliographystyle{IEEEtran}
\bibliography{ref}
\begin{IEEEbiography}[{\includegraphics[width=1in,height=1in,clip,keepaspectratio]{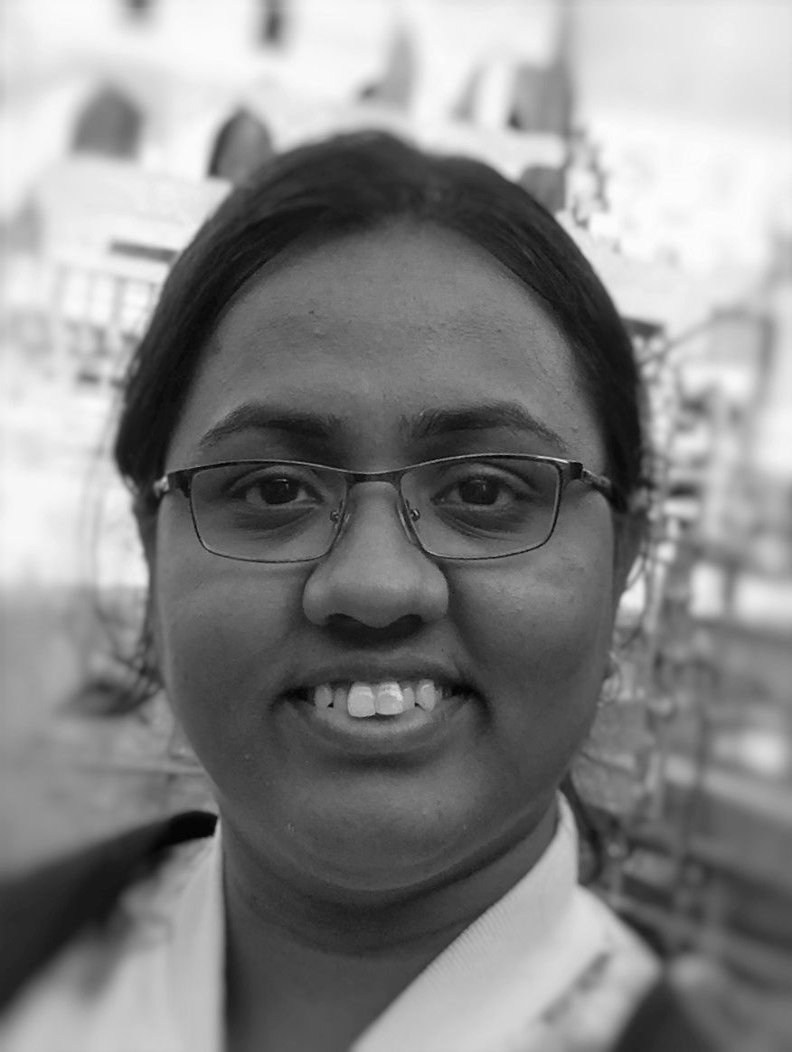}}]
{Soumi Chattopadhyay} (S'14, M'19) received her Ph.D. from Indian Statistical Institute in 2019. Currently, she is an Assistant Professor in Indian Institute of Information Technology Guwahati, India. Her research interests include Distributed and Services Computing, Artificial Intelligence, Formal Languages, Logic and Reasoning.
\end{IEEEbiography}
\begin{IEEEbiography}[{\includegraphics[width=1in,height=1in,clip,keepaspectratio]{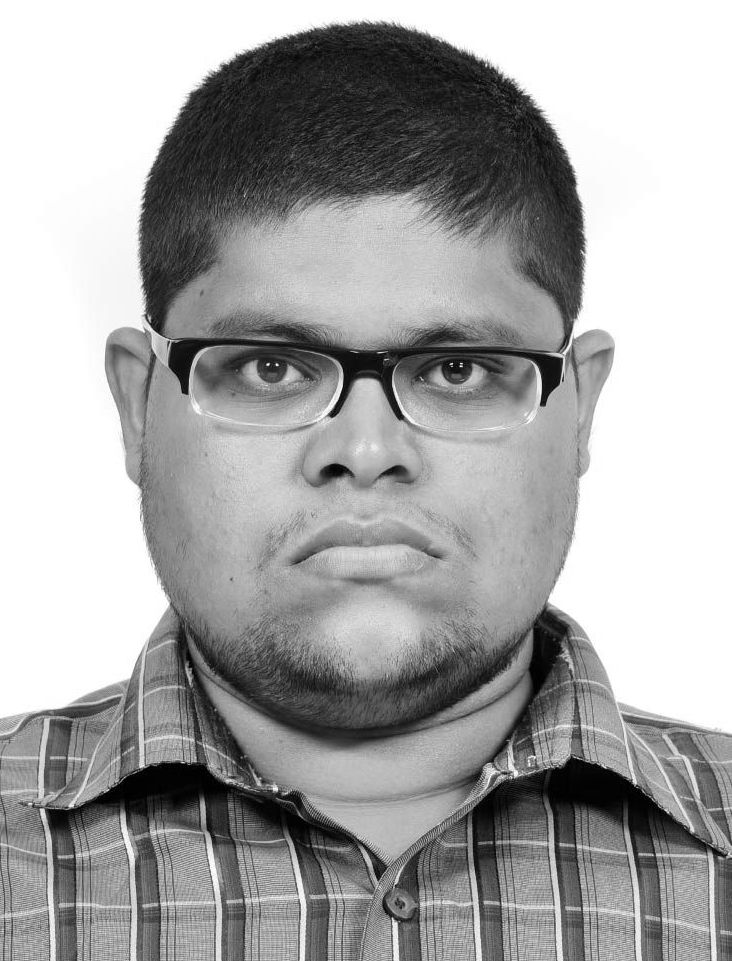}}]
{Chandranath Adak} (S'13, M'20) received his Ph.D. in Analytics from University of Technology Sydney, Australia in 2019. Currently, he is an Assistant Professor in Centre for Data Science, JIS Institute of Advanced Studies and Research, India. His research interests include Computer Vision and Machine Learning-related subjects.
\end{IEEEbiography}
\begin{IEEEbiography}[{\includegraphics[width=1in,height=1in,clip
,keepaspectratio]{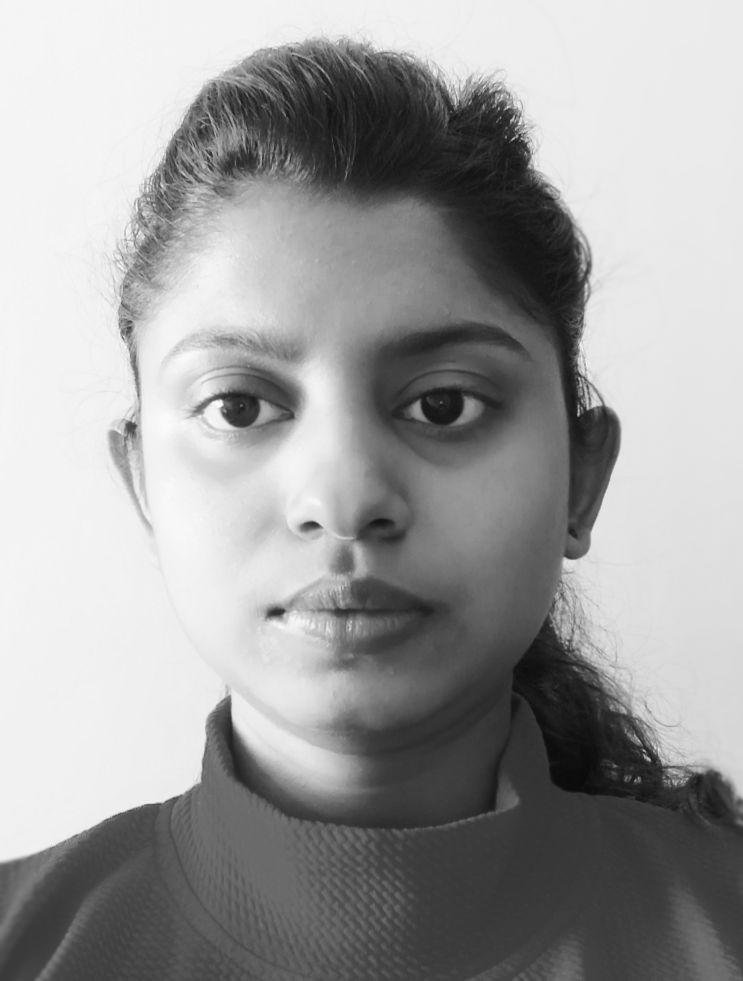}}]
{Ranjana Roy Chowdhury} received her M.Tech. in CSE from Indian Institute of Information Technology Guwahati, India in 2019. Currently, she is pursuing her Ph.D. from Indian Institute of Technology, Ropar.
Her research interests include Services Computing, Machine Learning-related subjects.
\end{IEEEbiography}
\end{document}